\documentclass{article}
\usepackage[final, nonatbib]{neurips_2024}
\usepackage[utf8]{inputenc} %
\usepackage[T1]{fontenc}    %
\usepackage[breaklinks,colorlinks,pdfa]{hyperref}
\hypersetup{
    pdftitle = {TrAct: Making First-layer Pre-Activations Trainable},
    pdflang = {en-US},
    urlcolor=blue!60!black,
    citecolor=blue!60!black,
    linkcolor=blue!60!black,
}
\usepackage{url}            %
\usepackage{booktabs}       %
\usepackage{amsfonts}       %
\usepackage{nicefrac}       %
\usepackage{microtype}      %
\usepackage{xcolor}         %
\usepackage{amsmath}
\usepackage{amssymb}
\usepackage{mathtools}
\usepackage{amsthm}
\usepackage{wrapfig}
\usepackage{tikz}
\usetikzlibrary{matrix, positioning, decorations.pathreplacing}
\usepackage{colortbl}
\usepackage[frozencache]{minted}
\definecolor{almond}{rgb}{0.94, 0.87, 0.8}

\usepackage[size=small]{caption}

\usepackage[backend=biber, style=ieee, citestyle=ieee, maxbibnames=15]{biblatex}
\AtBeginBibliography{\small}

\theoremstyle{plain}
\newtheorem{lemma}{Lemma}
\theoremstyle{remark}

\title{TrAct: Making First-layer Pre-Activations Trainable}

\author{
  Felix Petersen\\
  Stanford University\\
  {\small\texttt{mail@felix-petersen.de}}\\
  \And
  Christian Borgelt\\
  University of Salzburg\\
  {\small\texttt{christian@borgelt.net}}\\
  \And
  Stefano Ermon\\
  Stanford University\\
  {\small\texttt{ermon@cs.stanford.edu}}\\
}

\begin{document}
\maketitle

\begin{abstract}

We consider the training of the first layer of vision models and notice the clear relationship between pixel values and gradient update magnitudes: 
the gradients arriving at the weights of a first layer are by definition directly proportional to (normalized) input pixel values. 
Thus, an image with low contrast has a smaller impact on learning than an image with higher contrast, and a very bright or very dark image has a stronger impact on the weights than an image with moderate brightness.
In this work, we propose performing gradient descent on the embeddings produced by the first layer of the model.
However, switching to discrete inputs with an embedding layer is not a reasonable option for vision models.
Thus, we propose the conceptual procedure of (i) a gradient descent step on first layer activations to construct an activation proposal, and (ii) finding the optimal weights of the first layer, i.e., those weights which minimize the squared distance to the activation proposal. 
We provide a  closed form solution of the procedure and adjust it for robust stochastic training while computing everything efficiently. 
Empirically, we find that TrAct (Training Activations) speeds up training by factors between 1.25$\times$ and 4$\times$ while requiring only a small computational overhead. 
We demonstrate the utility of TrAct with different optimizers for a range of different vision models including convolutional and transformer architectures.
\end{abstract}

\section{Introduction}

We consider the learning of first-layer embeddings / pre-activations in vision models, and in particular learning the weights with which the input images are transformed in order to obtain these embeddings. 
In gradient descent, the updates to first-layer weights are directly proportional to the (normalized) pixel values of the input images. 
As a consequence (assuming that input images are standardized), high contrast, very dark, or very bright images have a greater impact on the trained first-layer weights, while low contrast images with medium brightness have only smaller impact on training. %

\begin{figure}[t]
\def\ourscale{0.1185}\setlength{\tabcolsep}{9pt}
\begin{tabular}{@{}ccc@{}}
Language&
Vision&
TrAct\\[.4em]
\fbox{\includegraphics[trim={482mm 153mm 115mm 153mm},clip,scale=.14]{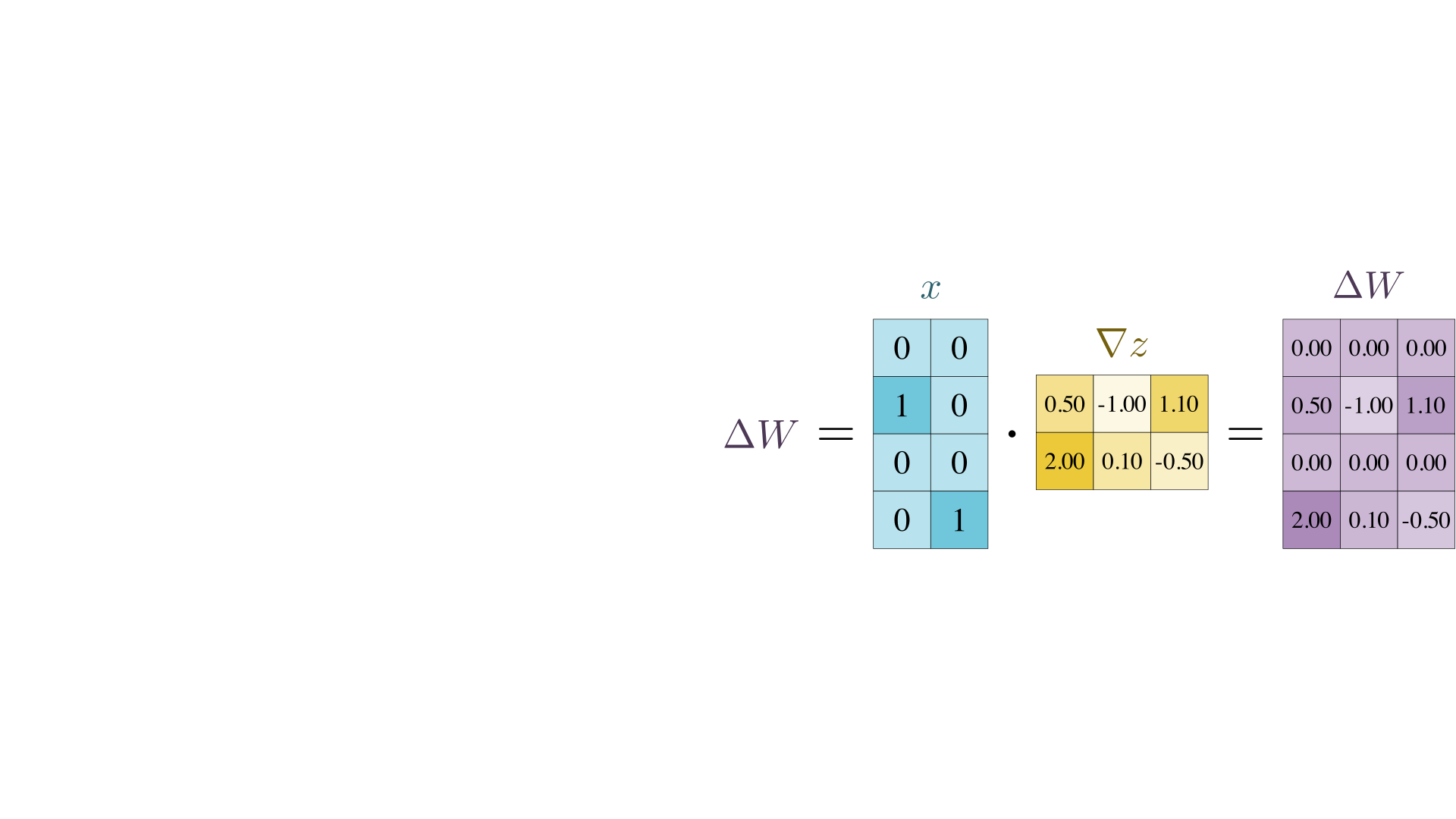}\raisebox{.81em}{$~=~$}\includegraphics[trim={596mm 152.5mm 0mm 153.5mm},clip,scale=.14]{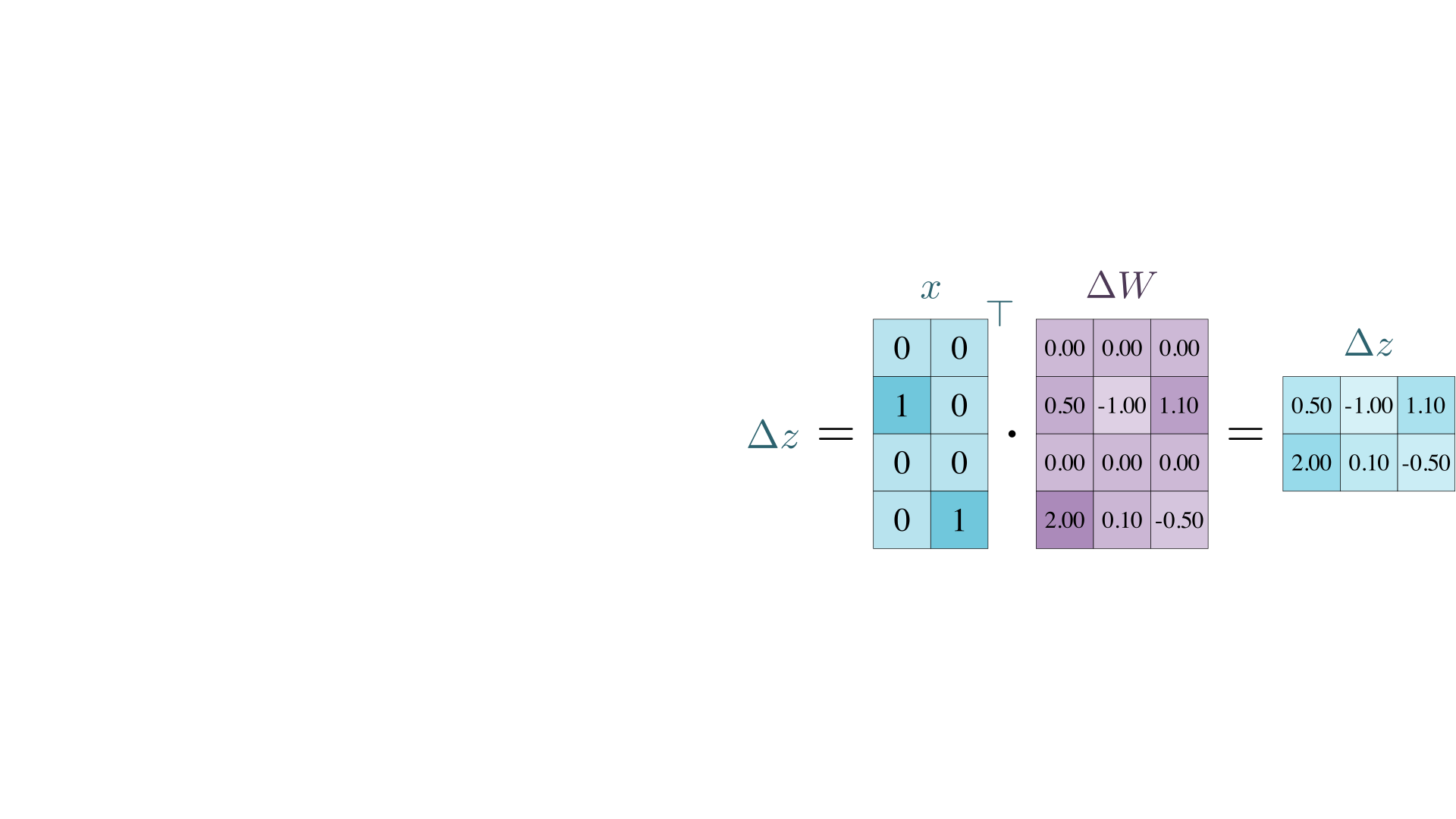}} 
&
\fbox{\includegraphics[trim={482mm 153mm 115mm 153mm},clip,scale=.14]{images/MatrixComparison0001}\raisebox{.81em}{$~\neq~$}\includegraphics[trim={596mm 152.5mm 0mm 153.5mm},clip,scale=.14]{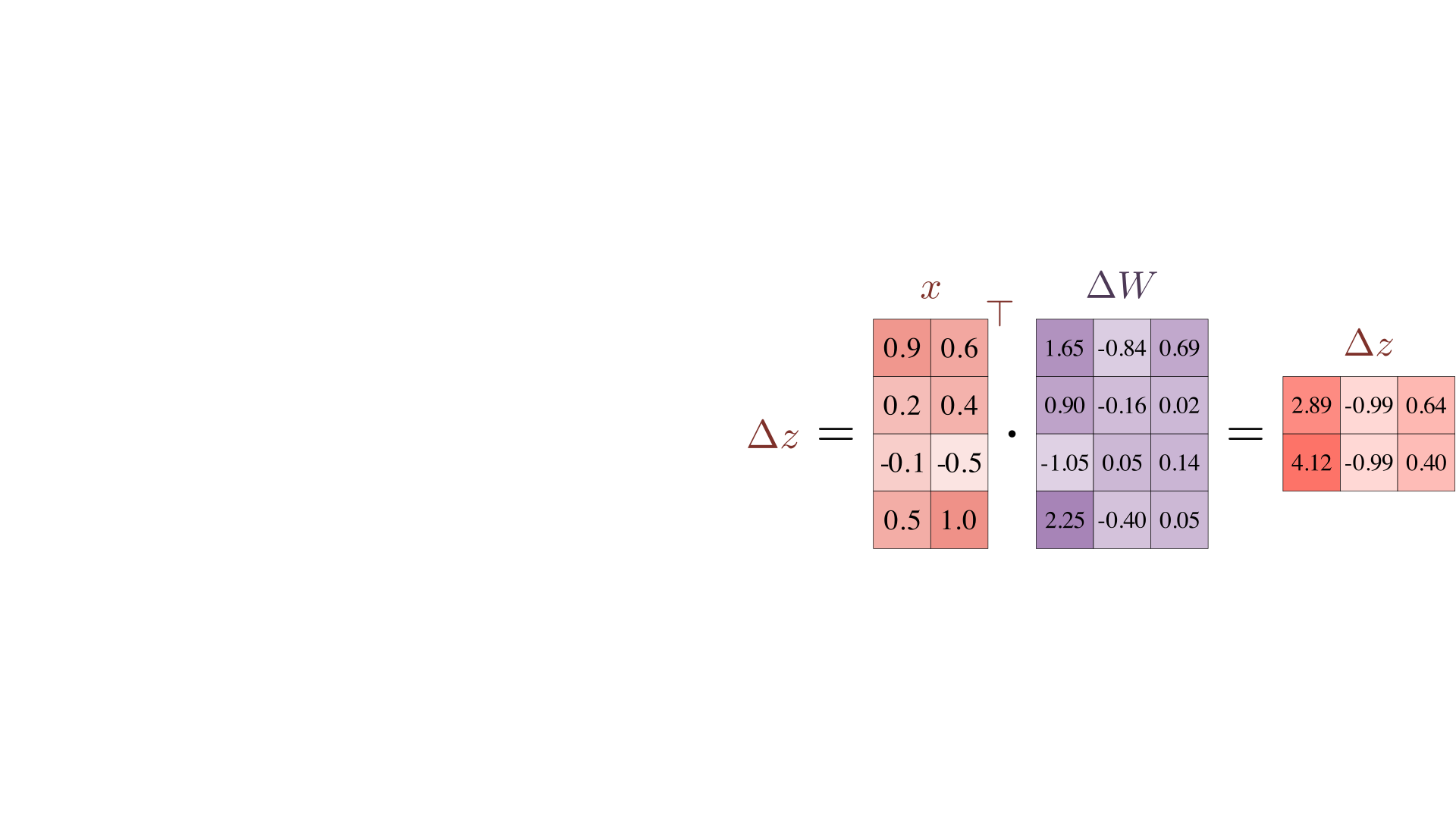}} 
&
\fbox{\includegraphics[trim={482mm 153mm 115mm 153mm},clip,scale=.14]{images/MatrixComparison0001}\raisebox{.81em}{$~\approx~$}\includegraphics[trim={596mm 152.5mm 0mm 153.5mm},clip,scale=.14]{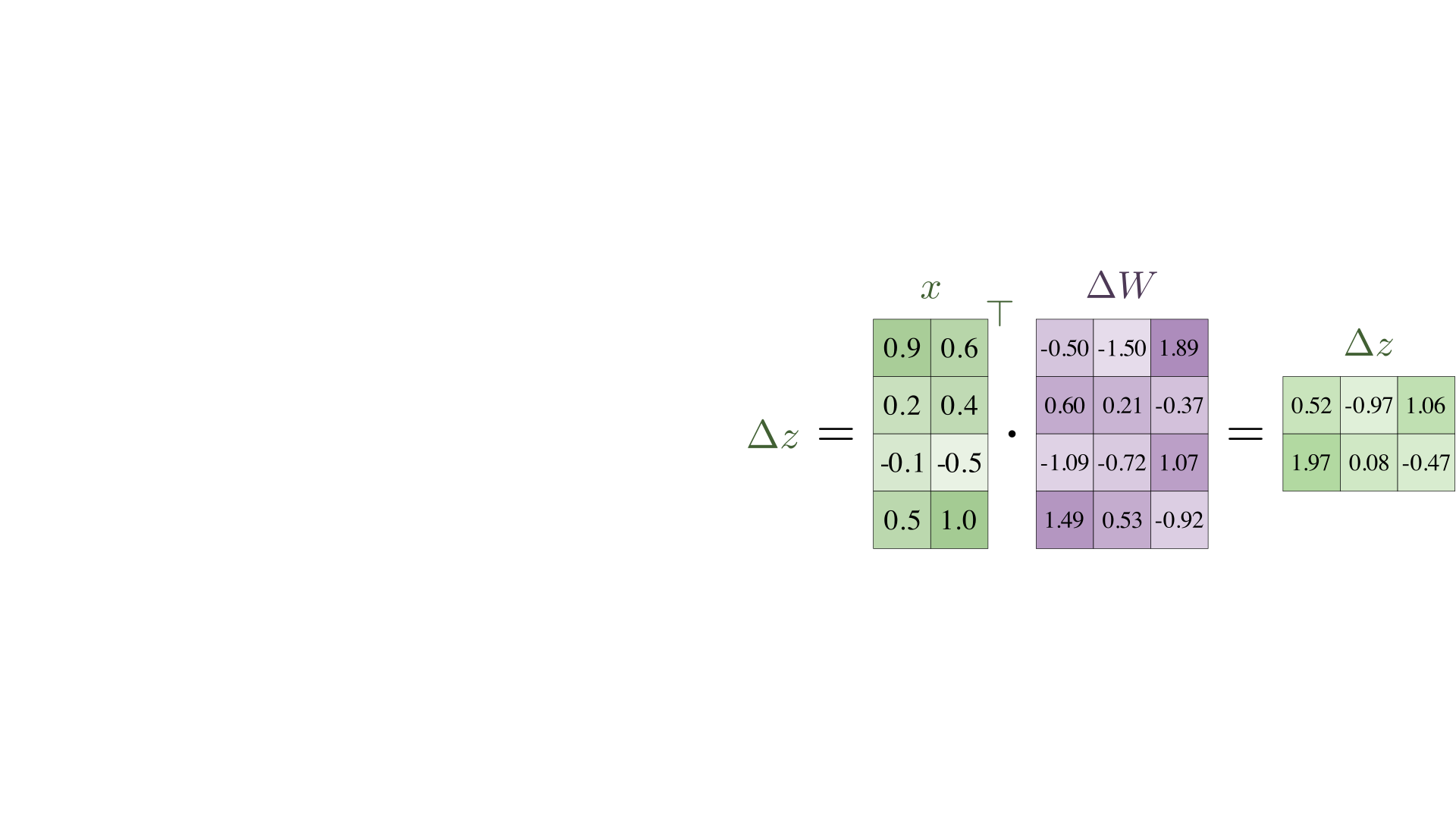}}
\\[.7em]
\includegraphics[trim={335mm 117mm 0mm 117mm},clip,scale=\ourscale]{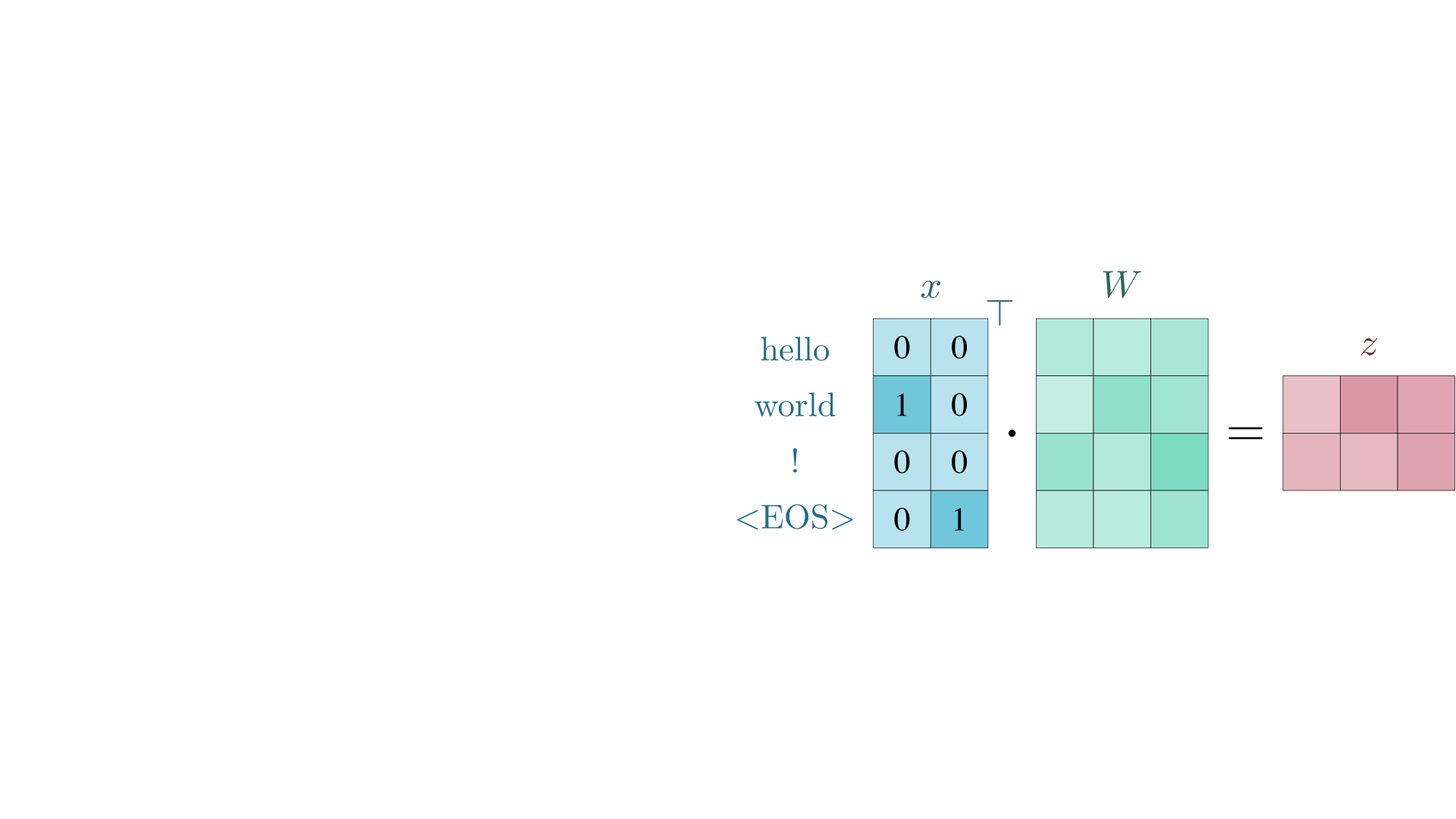} &
\includegraphics[trim={330mm 117mm 0mm 117mm},clip,scale=\ourscale]{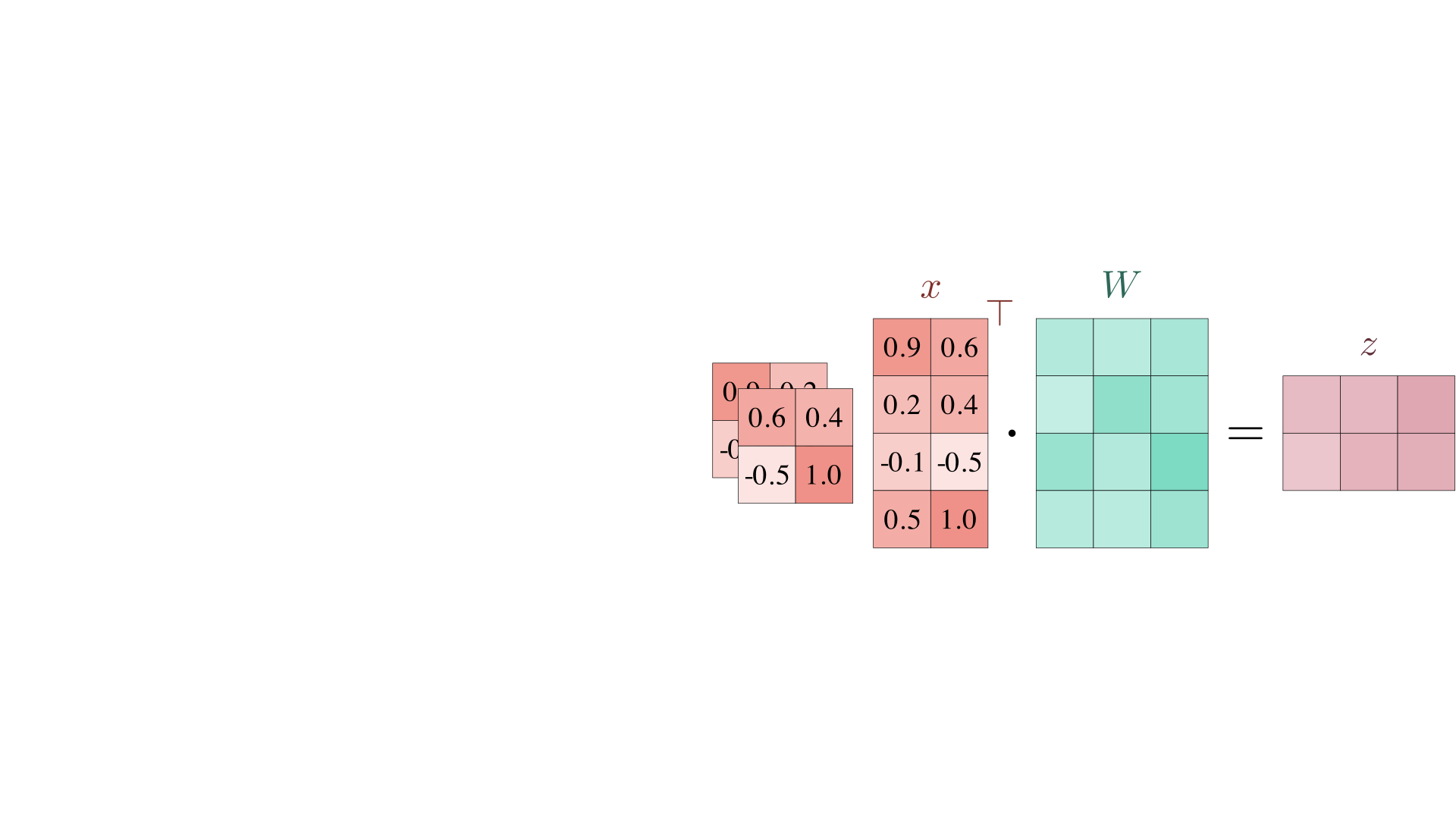} &
\includegraphics[trim={330mm 117mm 0mm 117mm},clip,scale=\ourscale]{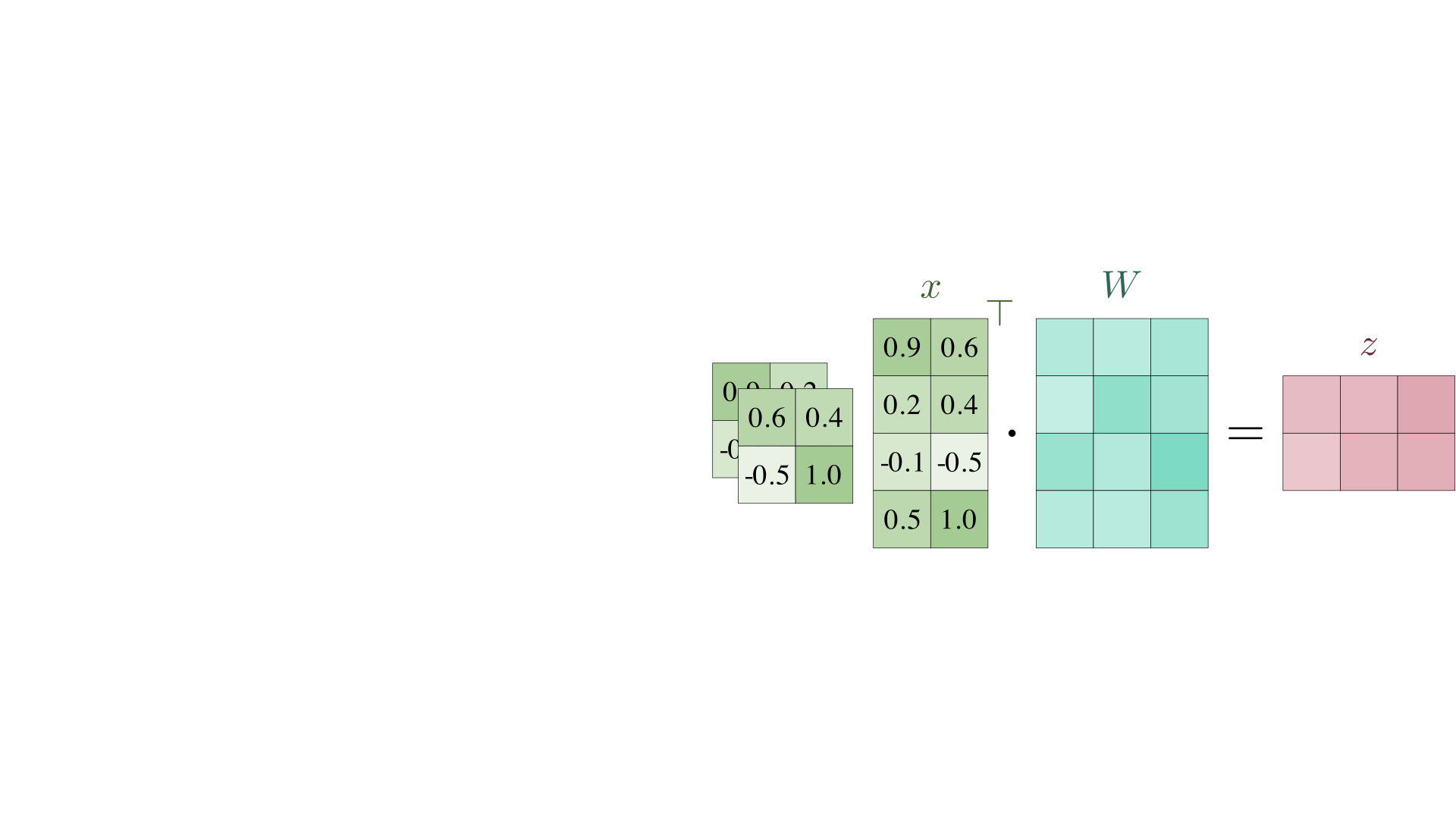} \\
\includegraphics[trim={335mm 117mm 0mm 117mm},clip,scale=\ourscale]{images/MatrixComparison0001} &
\includegraphics[trim={330mm 117mm 0mm 117mm},clip,scale=\ourscale]{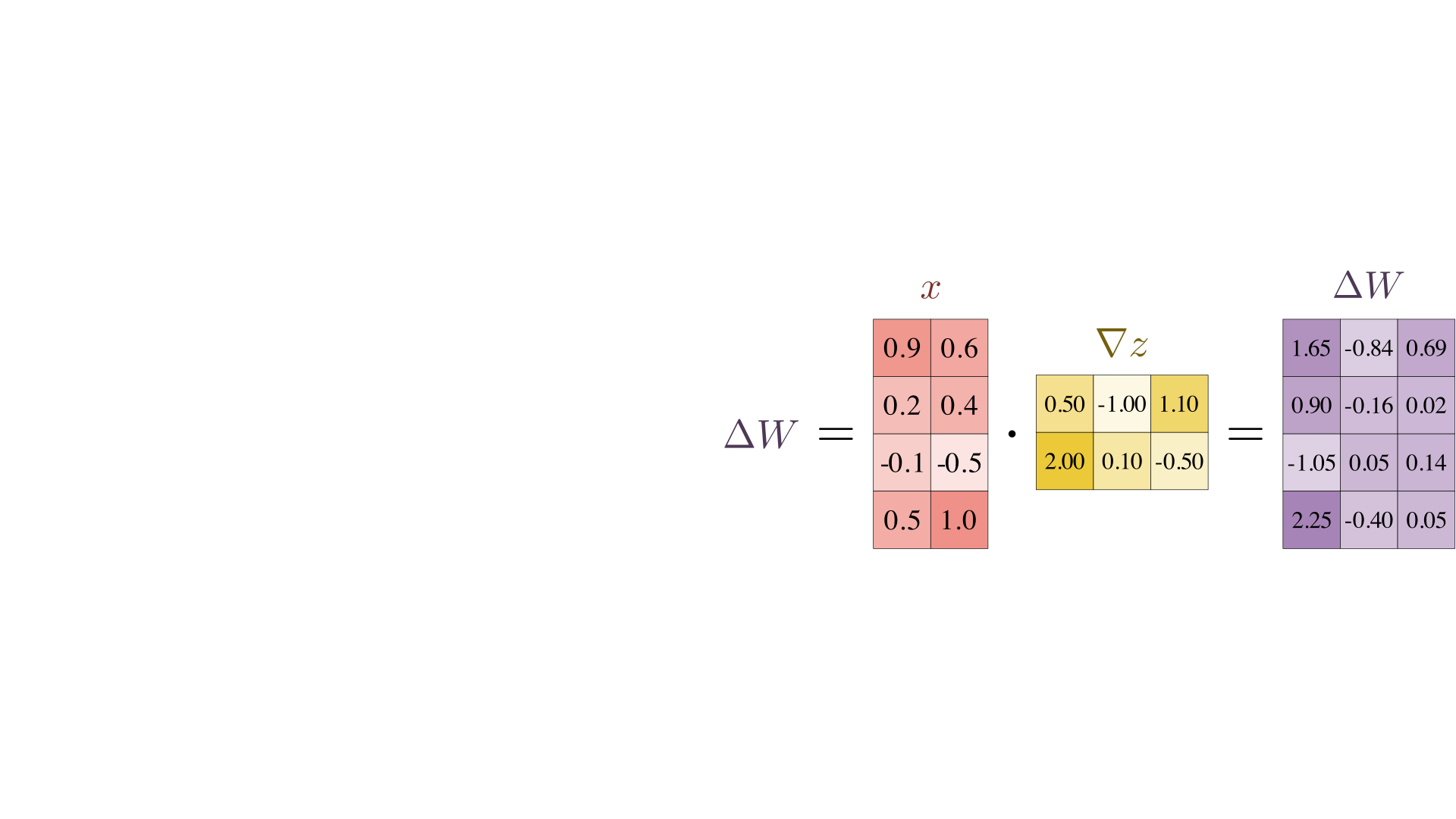} &
\hspace{-.25em}\includegraphics[trim={154mm 117mm 105mm 117mm},clip,scale=\ourscale]{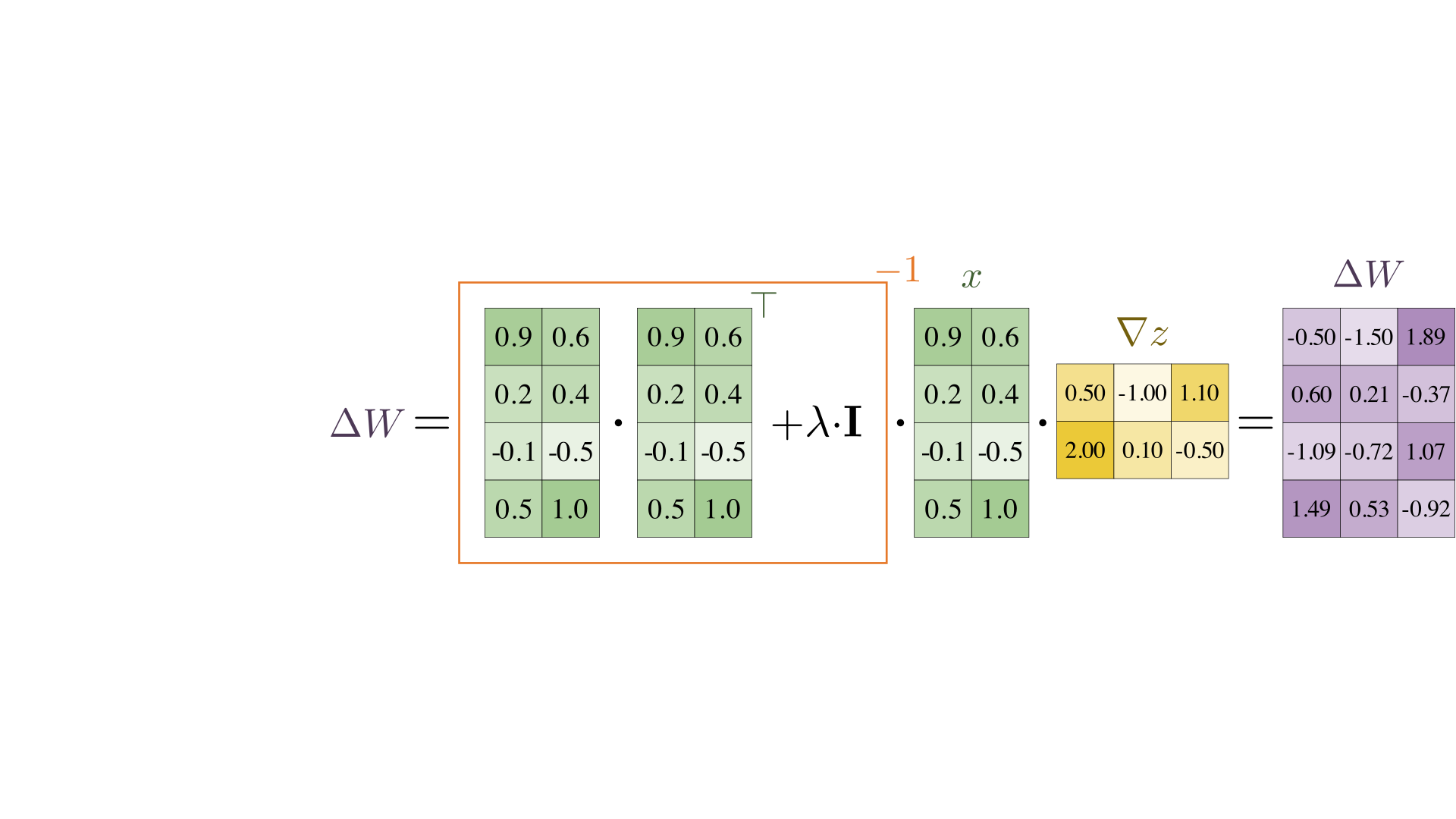} \\
\includegraphics[trim={335mm 117mm 0mm 117mm},clip,scale=\ourscale]{images/MatrixComparison0002} &
\includegraphics[trim={330mm 117mm 0mm 117mm},clip,scale=\ourscale]{images/MatrixComparison0005} &
\includegraphics[trim={330mm 117mm 0mm 117mm},clip,scale=\ourscale]{images/MatrixComparison0008} \\
\end{tabular}
\caption{
    TrAct learns the first layer of a vision model but with the training dynamics of an embedding layer.
    We~illustrate this in an example with two 4-dimensional inputs $x$, a weight matrix $W$ of size $4\times 3$, and resulting pre-activations $z$ of size $2\times3$.
    For language models (left), the input $x$ is two tokens from a dictionary of size~4.
    For vision models (center + right), the input $x$ is two patches of the image, each totaling 4 pixels.
    During backpropagation, we obtain the gradient wrt.\ our pre-activations $\nabla z$, from which the gradient and update to the weights $W$ is computed ($\Delta W$).
    The resulting update to the pre-activations $\Delta z$ equals $x^\top\cdot \Delta W$. 
    For language models (left), $\Delta z=\nabla z$, i.e., the training dynamics of the embeddings layer corresponds to updating the embeddings directly wrt.\ the gradient.
    Specifically, the update in a language model, for a token identifier $i$, is $W_i \leftarrow W_i - \eta \cdot \nabla_{z} \mathcal{L}(z)$ where $z = W_i$ is the activation of the first layer and at the same time the $i$th row of the embedding (weight) matrix $W$.
    Equivalently, we can write $z \leftarrow z - \eta \cdot \nabla_{z} \mathcal{L}(z)$.
    However, in vision models (center), the update $\Delta z$ strongly deviates from the respective gradients $\nabla z$.
    TrAct corrects for this by adjusting $\Delta W$ via a corrective term $(x \cdot x^\top + \lambda\cdot I)^{-1}$ (orange box), such that the update to $z$ closely approximates $\nabla z$.
\label{fig:overview}
}
\end{figure}

While, in the past, mainly transformations of the input images, especially various forms of normalization have been considered, either as a preprocessing step or as part of the neural network architecture, our approach targets the training process directly without modifying the model architecture or any preprocessing. 
The goal of our approach is to achieve a training behavior that is equivalent to training the pre-activations or embedding values themselves.
For example, in language models~\cite{vaswani2017attention}, the first layer is an ``Embedding'' layer that maps a token id to an embedding vector (via a lookup).
When training language models, this embedding vector is trained directly, i.e., the update to the embedding directly corresponds to the gradient of the pre-activation of the first layer.
As discussed above, this is not the case in vision models as, here, the updates to the first-layer weight matrix correspond to the outer product between the input pixel values and the gradient of the pre-activation of the first layer.
Bridging this gap between the ``Embedding'' layer in language models, and ``Conv2D'' / ``Linear'' / ``Dense'' layers in vision models, we propose a novel technique for training the pre-activations of the latter, effectively mimicking training behavior of the ``Embedding'' layer in language models.
As vision models rely on pixel values rather than tokens, and any discretization of image patches, e.g., via clustering is not a reasonable option, we approach the problem via a modification of the gradient (and therefore a modification of the training behavior) without modifying the original model architectures.
We illustratively compare the updates in language and vision models and demonstrate the modification that TrAct introduces in Figure~\ref{fig:overview}. 

The proposed method is general and applicable to a variety of vision model architecture types, from convolutional to vision transformer models.
In a wide range of experiments, we demonstrate the utility of the proposed approach, effectively speeding up training by factors ranging from $1.25\times$ to $4\times$, or, within a given training budget, improving model performance consistently.
The approach requires only one hyperparameter $\lambda$, which is easy to select, and our default value works consistently well across all $50$ considered model architecture + data set + optimizer settings.

\noindent
The remainder of this paper is organized as follows: 
in Section~\ref{sec:related-work}, we introduce related work,
in Section~\ref{sec:method}, we introduce and derive TrAct from a theoretical perspective, and in Section~\ref{sec:implementation} we discuss implementation considerations of TrAct.
In Section~\ref{sec:experiments}, we empirically evaluate our method in a variety of experiments, spanning a range of models, data sets, and training strategies, including an analysis of the mild behavior of the hyperparameter, an ablation study, and a runtime analysis.
We conclude the paper with a discussion in Section~\ref{sec:conclusion}.
\emph{The code is publicly available at \href{https://github.com/Felix-Petersen/tract}{github.com/Felix-Petersen/tract}.}

\section{Related Work}
\label{sec:related-work}

It is not surprising that the performance of image classification and object recognition models depends heavily on the quality of the input images, especially on their brightness range and contrast.
For example, image augmentation techniques generate modified versions of the original images as additional training examples. 
Some of these techniques work by geometric transformations (rotation, mirroring, cropping), others by adding noise, changing contrast or modifying the image in the color space~\cite{Shorten_and_Khoshgoftaar_2019}. 
In the area of vision transformers~\cite{Dosovitskiy_et_al_2020,Touvron_et_al_2021} so-called 3-augmentation (Gaussian blur, reduction to grayscale, and solarization) has been shown to be essential to performance~\cite{Touvron_et_al_2023}. 
Augmentation approaches are similar to image enhancement as a preprocessing step, because they generate possibly enhanced versions of the images as additional training examples, even though they leave the original images unchanged, which are also still used as training examples.%

Another direction related to the problem we deal with in this paper are various normalizations and standardizations, starting with the most common one of standardizing the data to mean~0 and standard
deviation~1 (over the training set), and continuing through batch normalization \cite{Ioffe_and_Szegedy_2015},
weight normalization~\cite{Salimans_and_Kingma_2016}, layer normalization~\cite{Ba_et_al_2016}, which are usually applied not just for the first layer, but throughout the network, and in particular patch-wise normalization of the input images~\cite{Kumar_et_al_2023}, which we will draw on for comparisons.
We note that, e.g., Dual PatchNorm~\cite{Kumar_et_al_2023}, in contrast to our approach, modifies the actual model architecture, but not the gradient backpropagation procedure.%

However, none of these approaches directly addresses the actual concern that weight changes in the first layer are proportional to the inputs, but instead only modify the inputs and architectures to make training easier or faster. 
In contrast to these approaches, we address the training problem itself and propose a different way of optimizing first-layer weights for unchanged inputs.
Of course, this does not mean that input enhancement techniques are superfluous with our method, but only that additional performance
gains can be obtained by including TrAct during training.

In the context of deviating from standard gradient descent-based optimization~\cite{dangel2023thesis}, there are different lines of work in the space of second-order optimization~\cite{wright2006numerical}, e.g., K-FAC~\cite{martens2015optimizing}, ViViT~\cite{dangel2022vivit}, ISAAC~\cite{Petersen_et_al_2023}, Backpack~\cite{dangel2020backpack}, and Newton Losses~\cite{petersen2024newton}, which have inspired our methodology for modifying the gradient computation.
In particular, the proposed approach integrates second-order ideas for solving a (later introduced) sub--optimization-problem in closed-form~\cite{hoerl1970ridge}, and has similarities to a special case of ISAAC~\cite{Petersen_et_al_2023}.%

\section{Method}
\label{sec:method}

First, let us consider a regular gradient descent of a vision model.
Let $z = f(x; W)$ be the first layer embeddings excluding an activation function and $W$ be the weights of this first layer, i.e., for a fully-connected layer $f(x; W) = W\cdot x$. 
Here, we have $x\in\mathbb{R}^{n\times b}$, $z\in\mathbb{R}^{m\times b}$, and $W\in\mathbb{R}^{m\times n}$ for a batch size of $b$.
We remark that our input $x$ may be unfolded, supporting convolutional and vision transformer networks.
Further, let $\hat y = g(\hat z; \theta_{\setminus W}) = g(f(x; W); \theta_{\setminus W})$ be the prediction of the entire model.
Moreover, let $\mathcal{L}(\hat y, y)$ be the loss function for a label $y$ and wlog.~let us assume it is an averaging loss (i.e., reduction over batch dimension via mean).
During backpropagation, the gradient of the loss wrt.~$z$, i.e.,  $\nabla_{\!z}\, \mathcal{L}(g(z; \theta_{\setminus W}), y)$~~or~~$\nabla_{\!z} \mathcal{L}(z)$ for short, will be computed.
Conventionally, the gradient wrt.~$W$, i.e., $\nabla_{W} \mathcal{L}(g(f(x; W); \theta_{\setminus W}), y)$~~or~~$\nabla_{W} \mathcal{L}(W)$ for short, is computed during backpropagation as 
\vspace*{-1.5ex}
\begin{align}
\nabla_{W} \mathcal{L}(W)
= \nabla_{\!z} \mathcal{L}(z) \cdot x^\top\,,
\label{eq:grad-w}
\end{align}
leading to the gradient descent update step of 
\begin{align}
W \leftarrow W - \eta \cdot
\nabla_{W} \mathcal{L}(W)\,.
\end{align}
Equation~\ref{eq:grad-w} clearly shows the direct proportionality between the gradient wrt.~the first layer weights and the input (magnitudes), showing that larger input magnitudes produce proportionally larger changes in first layer network weights. We remark that a corresponding relationships also holds in later layers of the neural network, but emphasize that, in later layers, the relationship shows a proportionality to activation magnitude, which is desirable. 

To resolve this dependency on the inputs and make training more efficient, we propose to conceptually optimize in the space of first layer embeddings $z$.
In particular, we could perform a gradient descent step on $z$, i.e.,\vspace{-.5em}
\begin{align}
z^\star \leftarrow z - \eta \cdot b \cdot \nabla_{\!z} \mathcal{L} (z)\,.
\end{align}
Here, $b$ is a multiplier because $\mathcal{L} (z)$ is (per convention) the empirical expectation over the\,batch\,dim.

However, now, $z^\star$ depends on the inputs and is not part of the actual model parameters.
We can resolve this problem by determining how to update $W$ such that $f(x; W)$ is as close to $z^\star$ as possible.
Conceptually, we compute the optimal update $\Delta W^\star$ by solving the optimization problem 
\begin{equation}
\arg\min_{\Delta W} ~ \| z^\star - (W + \Delta W) \cdot x \|_2^2 \\
\qquad\mathrm{subject~to}\quad\| \Delta W \|_2 \leq \epsilon
\label{eq:original-problem}
\end{equation}
where we (1)~want to minimize the distance between $z^\star$ the embeddings implied by the change of $W$ by $\Delta W$,
and (2)~want to keep the change $\Delta W$ small.

We enforce that weight matrix changes $\Delta W$ are small ($\| \Delta W \|_2{\leq}\epsilon$) by taking the Lagrangian of the problem, i.e., %
\begin{align}
    \arg\min_{\Delta W} ~ \| z^\star - (W + \Delta W) \cdot x \|_2^2 \,+\, \lambda b \cdot \| \Delta W \|_2^2
    \label{eq:lagrangian-problem}
\end{align}
with a heuristically selected Lagrangian multiplier $\lambda\cdot b$ (parameterized with $b$ because the first part is also proportional to $b$). 
We simplify Equation~\ref{eq:lagrangian-problem} to 
\begin{equation}
    \arg\min_{\Delta W} ~\| - \eta b \cdot \nabla_{\!z} \mathcal{L} (z) - \Delta W \cdot x \|_2^2 \,+\, \lambda b \cdot \| \Delta W \|_2^2
\end{equation}
and ease the presentation by considering it from a row-wise perspective, i.e., for $\Delta W_i\in\mathbb{R}^{1\times n}$:
\begin{equation}
    \arg\min_{\Delta W_i} ~\| - \eta b \cdot \nabla_{\!z_i} \mathcal{L} (z) - \Delta W_i \cdot x \|_2^2 \,+\, \lambda b \cdot \| \Delta W_i \|_2^2\,.
    \label{eq:problem-wi}
\end{equation}
The problem is separable into a row-wise perspective because the norm ($\|\cdot\|_2^2$) is the squared Frobenius norm and the rows have independent solutions.

In the following, we provide a closed-form solution for optimization problem~\eqref{eq:problem-wi}, which is related to~\cite{hoerl1970ridge, calvetti2004tikhonov}. 
\begin{lemma}
The solution $\Delta W_i^\star$ of Equation~\ref{eq:problem-wi} is%
\begin{align}
    \Delta W_i^\star
    &= -\,\eta \cdot \nabla_{\!z_i} \mathcal{L}(z)  \cdot x^\top \cdot \left(\frac{xx^\top\!}{b} + \lambda\cdot I_n\right)^{-1} \!.
\end{align}
\textit{Proof deferred to Supplementary Material~\ref{sm:theory}.}
\end{lemma}

\noindent
Extending the solution to $\Delta W$, we have %
\begin{align}
    \Delta W^\star 
    &= -\,\eta  \cdot \nabla_{\!z} \mathcal{L}(z)  \cdot x^\top \cdot \left(\frac{xx^\top\!}{b} + \lambda\cdot I_n\right)^{-1}
    \label{eq:combining-rows-from-lemma-1}
\end{align}
and can accordingly use it for an update step for $W$, i.e., %
\begin{align}
    W \leftarrow W+\Delta W^\star \qquad\text{or}\qquad
    W \leftarrow W-\eta  \cdot \nabla_{\!z} \mathcal{L}(z)  \cdot x^\top \cdot \left(\frac{xx^\top\!}{b} + \lambda\cdot I_n\right)^{\!\!-1}\!\!.
    \label{eq:w-update-full}
\end{align}
The update in Equation~\ref{eq:w-update-full} directly inserts the solution of the problem formulated in Equation~\ref{eq:original-problem}.
This computation is efficient as it only requires inversion of an $n\times n$ matrix, where $n$ in the case of convolutions correspond to $3$ (RGB) times the squared first layer's kernel size, and for vision transformers corresponds to the number of pixels per patch. The values of $n$ typically range from $n=27$ (CIFAR ResNet) to $n=768$ (ImageNet large-scale vision transformer).

\begin{lemma}
    Using TrAct does not change the set of possible convergence points compared to vanilla (full batch) gradient descent.
    Herein, we use the standard definition of convergence points as those points where no update is performed because the gradient is zero.
\end{lemma}
\noindent\textit{Proof sketch:}
    First, we remark that only the training of the first layer is affected by TrAct.
    To show the statement, we show that (i) a zero gradient for GD implies that TrAct also performs no update and that (ii) TrAct performing no update implies zero gradients for GD.
    \textit{Proof deferred to SM~\ref{sm:theory}.}\\[.5em]
The statement formalizes that TrAct does not change the set of attainable models, but instead only affects the behavior of the optimization itself.

For an illustration of how TrAct affects updates to $W$ and $z$, with a comparison to language models and conventional vision models, see Figure~\ref{fig:overview}.

\subsection{Implementation Considerations}
\label{sec:implementation}

To implement the proposed update in Equation~\ref{eq:w-update-full} in modern automatic differentiation frameworks~\cite{paszke2019pytorch, jax2018github}, we can make use of a custom backward or backpropagation for the first layer. 

\begin{figure}[h]
\vspace{-.25em}
\begin{minipage}[t]{0.39\linewidth}
Standard gradient for the weights:
\begin{align}
    \nabla_W \leftarrow \nabla_{\!z} \mathcal{L}(z) \cdot x^\top
    \phantom{\hbox to0pt{$\Big)^{-1}$\hss}}
\end{align}
implemented via a backward function:
\begin{minted}[escapeinside=||,fontsize=\small,fontfamily=cmtt]{python}
  def backward(grad_z, x, W):
      grad_W = grad_z.T @ x
      return grad_W
\end{minted}
\end{minipage}\hfill
\begin{minipage}[t]{0.55\linewidth}
For TrAct, we perform an in-place replacement by:
\begin{align}
    \nabla_W \leftarrow \nabla_{\!z} \mathcal{L}(z)  \cdot x^\top \cdot  \Big({\textstyle\frac{xx^\top\!}{b}} + \lambda\cdot I_n\Big)^{-1}
\end{align}
i.e., we replace the backward of the first layer by:
\begin{minted}[escapeinside=||,fontsize=\small,fontfamily=cmtt]{python}
 def backward(grad_z, x, W, l=0.1):
     b, n = x.shape
     grad_W = grad_z.T @ x @ inverse(
              x.T @ x / b + l * eye(n))
     return grad_W
\end{minted}
\end{minipage}
\caption{\label{fig:implementation}Implementation of TrAct, 
where \textbf{\texttt{l}} corresponds to the hyperparameter $\lambda$.}\vspace{-1em}
\end{figure}

Details are shown in Figure~\ref{fig:implementation}.
This applies the TrAct update from Equation~\ref{eq:w-update-full} when using the SGD optimizer.
Moreover, extensions of the update corresponding to optimizers like ADAM~\cite{kingma2015adam} (including, e.g., momentum, learning rate scheduler, regularizations, etc.) can be attained by using a respective optimizer and pretending (towards the optimizer) that the TrAct update corresponds to the gradient. 
As it only requires a modification of the gradient computation of the first layer, the proposed method allows for easy adoption in existing code.
All other layers / weights ($\theta_{\setminus W}$) are trained conventionally without modification.
Convolutions can be easily expressed as a matrix multiplication via an unfolding of the input; accordingly, we unfold the inputs respectively in the case of convolution.
\\[-1em]

Moreover, we would like to show a second method of applying TrAct that is exactly equivalent.
Typically, we have some batch of data \textbf{\texttt{x}} and a first embedding layer \textbf{\texttt{embed}}, as well as a remaining network \textbf{\texttt{net}}, a \textbf{\texttt{loss}}, and targets \textbf{\texttt{gt}}.
In the following, we show how the forward and backward is usually written (left) and how it can be modified to incorporate TrAct (right):
\\[.5em]
\noindent 
\iftrue
\begin{minipage}[t]{0.44\textwidth}
\begin{minted}[escapeinside=||,fontsize=\small,fontfamily=cmtt]{python}
 z = embed(x)  # first layer pre-act
 y = net(z)   # remainder of the net
 loss(y, gt).backward()   # backprop
\end{minted}
\end{minipage}\hfill
\begin{minipage}[t]{0.52\textwidth}
\begin{minted}[escapeinside=||,fontsize=\small,fontfamily=cmtt]{python}
 z = embed(x @ inverse(x.T @ x/b+l*eye(n)))
 z.data = embed(x)  # overwrites the values
 # in z but leaves the gradient as before
 y = net(z)
 loss(y, gt).backward()
\end{minted}
\end{minipage}
\else
The forward and backward application is usually written as:
\begin{minted}[escapeinside=||,fontsize=\small,fontfamily=cmtt]{python}
   z = embed(x)   # first layer pre-activations
   y = net(z)        # remainder of the network
   loss(y, gt).backward()     # backpropagation
\end{minted}
Indeed, we can apply TrAct by modifying this to
\begin{minted}[escapeinside=||,fontsize=\small,fontfamily=cmtt]{python}
 z = embed(x @ inverse(x.T @ x / b + l * eye(n)))
 z.data = embed(x).stop_grad()  # overwrites the 
 # values in z but leaves the gradient as before
 y = net(z)
 loss(y, gt).backward()
\end{minted}
\fi
\\[.5em]
This modifies the input of \textbf{\texttt{embed}} for the gradient computation, but replaces the actual values propagated through the remaining network \textbf{\texttt{z.data}} by the original values, therefore not affecting downstream layers.
which modifies the input of \textbf{\texttt{embed}} for the gradient computation, but replaces the actual values propagated through the remaining network \textbf{\texttt{z.data}} by the original values, therefore not affecting downstream layers.
This illustrates interesting relationships: TrAct is minimally invasive, can be removed or included at any time without breaking the network, and does not have learnable parameters.
TrAct can be seen as in some sense related to normalizing / whitening / inverting the input for the purpose of gradient computation, but then switching the embeddings back to the original first layer embeddings / activations for propagation through the remainder of the network.%

We provide an easy-to-use wrapper module that can be applied to the first layer, and automatically provides the TrAct gradient computation replacement procedure.
For example, for PyTorch~\cite{paszke2019pytorch}, the \textbf{\texttt{TrAct}} module can be applied to {\texttt{nn.Linear}} and {\texttt{nn.Conv2d}} layers by wrapping them as 
\begin{minted}[escapeinside=||,fontsize=\normalsize,fontfamily=cmtt]{python}
            |\color{blue}{TrAct}|(nn.|{Linear}|(...))       |\color{blue}{TrAct}|(nn.|{Conv2d}|(...))
\end{minted}
and for existing implementations, we can apply TrAct, e.g., for vision transformers via:
\begin{minted}[escapeinside=||,fontsize=\normalsize,fontfamily=cmtt]{python}
           net.patch_embed.proj = |\color{blue}{TrAct}|(net.patch_embed.proj)
\end{minted}

\section{Experimental Evaluation}
\label{sec:experiments}

\subsection{CIFAR-10}
\label{sec:cifar-10}

\paragraph{Setup}
For the evaluation on the CIFAR-10 data set~\cite{krizhevsky2009cifar10}, we consider the ResNet-$18$~\cite{he2016deep} as well as a small ViT model.
We consider training from scratch as the method is particularly designed for this case.
We perform training for $100$, $200$, $400$, and $800$ epochs.
For the ResNet models, we use the Adam and SGD with momentum ($0.9$) optimizers, both with cosine learning rate schedules; learning rates, due to their significance, will be discussed alongside respective experiments.
Further, we use the standard softmax cross-entropy loss.
For the ViT, we use Adam with a cosine learning rate scheduler as well as a softmax cross-entropy loss with label smoothing ($0.1$).
The selected ViT\footnote{Based on \href{https://github.com/omihub777/ViT-CIFAR/}{github.com/omihub777/ViT-CIFAR}.} is particularly designed for effective training on CIFAR scales and has $7$ layers, $12$ heads, and hidden sizes of $384$.
Each model is trained with a batch size of $128$ on an Nvidia RTX 4090 GPU with PyTorch~\cite{paszke2019pytorch}.

As mentioned above, the learning rate is a significant factor in the evaluation.
Therefore, throughout this paper, to remove any bias towards the proposed method (and even give an advantage to the baseline), we utilize the optimal learning rate of the baseline also for the proposed method.
For the Adam optimizer, we consider a learning rate grid of $\{10^{-2}, 10^{-2.5}, 10^{-3}, 10^{-3.5}\}$;
for SGD with momentum, a learning rate grid of $\{0.1, 0.09, 0.08, 0.07\}$.
The optimal learning rate is determined for each number of epochs using regular training; in particular,
for Adam, we have 
$\{100{\to}10^{-2}, $
$200{\to}10^{-2}, \allowbreak400{\to}10^{-3}, \allowbreak800{\to}10^{-3}\}$, 
and, for SGD with momentum, we find that a learning rate of $0.08$ is optimal in each case.
For the ViT, we considered a learning rate grid of $\{$$10^{-3}, \allowbreak10^{-3.1}, \allowbreak10^{-3.2}, \allowbreak10^{-3.3}, \allowbreak10^{-3.4}, \allowbreak10^{-3.5}, \allowbreak10^{-3.6}, \allowbreak10^{-3.7}, \allowbreak10^{-3.8}, \allowbreak10^{-3.9}, \allowbreak10^{-4}$$\}$.
Here, the optimal learning rates (based on the baseline) are $\{100{\to}10^{-3}, \allowbreak200{\to}10^{-3.2}, \allowbreak400{\to}10^{-3.5}, \allowbreak800{\to}10^{-3.5}\}$.

\begin{figure*}[t]
    \includegraphics[width=.49\linewidth]{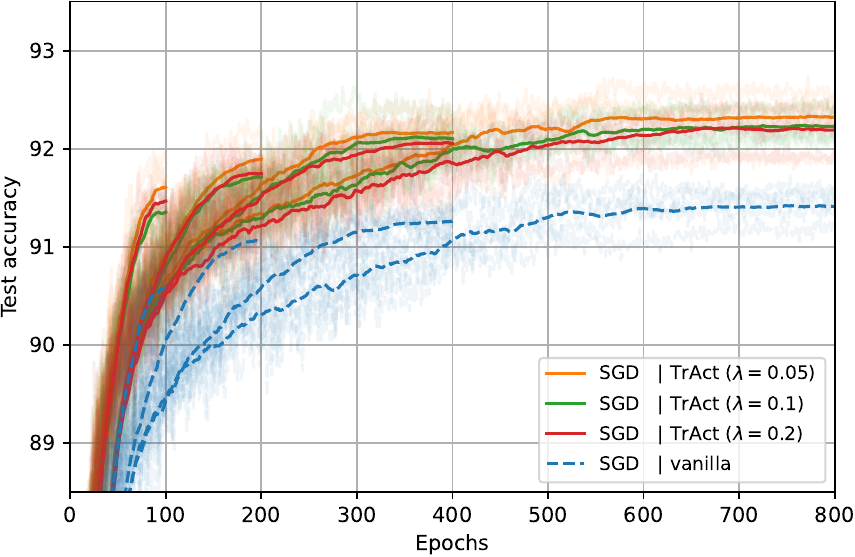}\hfill%
    \includegraphics[width=.49\linewidth]{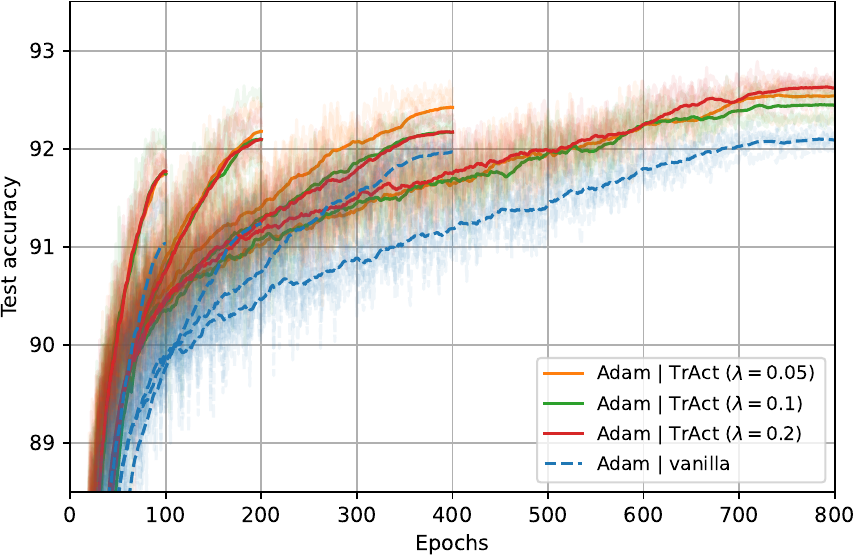}
    \caption{\label{fig:resnet-18-cifar-10}%
    Training a \textbf{ResNet-18} on \textbf{CIFAR-10}. 
    We train for $\{100,200,400,800\}$ epochs using a cosine learning rate schedule and with SGD (left) and Adam (right). 
    Learning rates have been selected as optimal for each baseline. 
    Averaged over 5 seeds. 
    TrAct (solid lines) consistently outperforms the baselines (dashed)---in many cases already with a quarter of the number of the epochs of the baseline.
    }
\end{figure*}

\vspace{-.75em}
\paragraph{Results}
In Figure~\ref{fig:resnet-18-cifar-10} we show the results for ResNet-18 trained on CIFAR-10.
We can observe that TrAct improves the test accuracy in every setting, in particular, for both optimizers, for all four numbers of epochs, and for all three choices of the hyperparameter $\lambda\in$
$\{0.05, 0.1, 0.2\}$.
Moreover, we can observe that, 
for SGD, the accuracy after 100 epochs is already better than for the baseline after 800 epochs. 
For Adam, we can see that TrAct after 100
epochs performs similar to the baseline after 400 epochs, and TrAct after 200 epochs performs similar to the baseline after 800 epochs.
Comparing the different choices of $\lambda$, $\lambda=0.05$ performs best in most cases.

\begin{wrapfigure}[14]{r}{0.5\linewidth}
    \centering
    \vspace{-1em}
    \includegraphics[width=\linewidth]{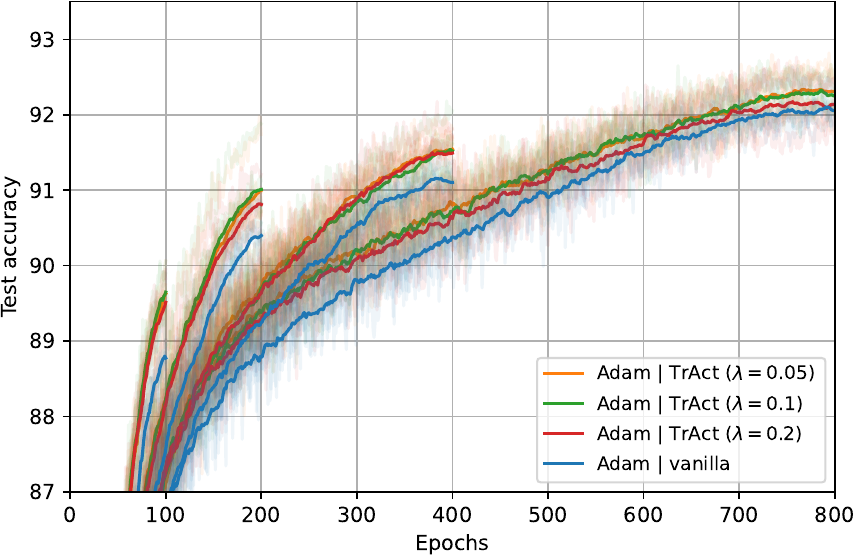}
    \vskip-0.5ex
    \caption{\label{fig:vit-cifar-10}Training a \textbf{ViT} on \textbf{CIFAR-10}. 
    We train for $\{100,200,400,800\}$ epochs using a cosine learning rate schedule and with Adam. 
    Learning rates have been selected as optimal for each baseline. 
    Avg.\ over 5 seeds.}
    \vskip-4em
\end{wrapfigure}

The results for the ViT model are displayed in
Figure~\ref{fig:vit-cifar-10}.
Again, we can observe that TrAct consistently outperforms the baselines for all $\lambda$. 
Further, we can observe that TrAct with 200 epochs performs comparable to the baseline with 400 epochs.
We emphasize that, again, the optimal learning rate has been selected based on the baseline.
Overall, here, $\lambda=0.1$ performed best.

\subsection{CIFAR-100}

\paragraph{Setup}
{For CIFAR-100, we consider two experimental settings.
First, we consider the training of $36$ different convolutional model architectures based on a strong and popular repository\footnote{Based on \href{https://github.com/weiaicunzai/pytorch-cifar100/}{github.com/weiaicunzai/pytorch-cifar100}.} for CIFAR-100.
We use the same {hyperparame-\parfillskip=0pt \par}

\begin{wraptable}[40]{r}{0.58\linewidth}
    \centering
    \setlength{\tabcolsep}{2.5pt}
    \footnotesize
    \begin{tabular}{l|cc|cc}
\toprule
                                                & \multicolumn{2}{c}{Baseline} & \multicolumn{2}{c}{TrAct ($\lambda{=}0.1$)} \\
Model                                           & Top-1 & Top-5 & Top-1 & Top-5 \\
\midrule
SqueezeNet~\cite{iandola2016squeezenet}         & $69.45\%$ & $91.09\%$ & $\mathbf{70.48}\%$ & $\mathbf{91.50}\%$ \\ %
MobileNet~\cite{howard2017mobilenets}           & $66.99\%$ & $88.95\%$ & $\mathbf{67.06}\%$ & $\mathbf{89.12}\%$ \\ %
MobileNetV2~\cite{sandler2018mobilenetv2}       & $67.76\%$ & $90.80\%$ & $\mathbf{67.89}\%$ & $\mathbf{90.91}\%$ \\ %
ShuffleNet~\cite{zhang2018shufflenet}           & $\mathbf{69.98}\%$ & $91.18\%$ & $       {69.97}\%$ & $\mathbf{91.45}\%$ \\ %
ShuffleNetV2~\cite{ma2018shufflenet}            & $69.31\%$ & $90.91\%$ & $\mathbf{69.88}\%$ & $\mathbf{91.02}\%$ \\ %
VGG-11~\cite{simonyan2014very}                  & $68.44\%$ & $88.02\%$ & $\mathbf{69.66}\%$ & $\mathbf{88.99}\%$ \\ %
VGG-13~\cite{simonyan2014very}                  & $71.96\%$ & $90.27\%$ & $\mathbf{72.98}\%$ & $\mathbf{90.78}\%$ \\ %
VGG-16~\cite{simonyan2014very}                  & $72.12\%$ & $89.81\%$ & $\mathbf{72.73}\%$ & $\mathbf{90.11}\%$ \\ %
VGG-19~\cite{simonyan2014very}                  & $71.13\%$ & $88.10\%$ & $\mathbf{71.45}\%$ & $\mathbf{88.42}\%$ \\ %
DenseNet121~\cite{huang2017densely}             & $78.93\%$ & $94.83\%$ & $\mathbf{79.55}\%$ & $\mathbf{94.92}\%$ \\ %
DenseNet161~\cite{huang2017densely}             & $79.95\%$ & $95.25\%$ & $\mathbf{80.47}\%$ & $\mathbf{95.37}\%$ \\ %
DenseNet201~\cite{huang2017densely}             & $79.39\%$ & $95.07\%$ & $\mathbf{79.94}\%$ & $\mathbf{95.17}\%$ \\ %
GoogLeNet~\cite{szegedy2014going}               & $76.85\%$ & $93.53\%$ & $\mathbf{77.18}\%$ & $\mathbf{93.86}\%$ \\ %
Inception-v3~\cite{szegedy2016rethinking}       & $\mathbf{79.40}\%$ & $94.94\%$ & $79.24\%$ & $\mathbf{95.04}\%$ \\ %
Inception-v4~\cite{szegedy2017inception}        & $\mathbf{77.32}\%$ & $93.80\%$ & $77.14\%$ & $\mathbf{93.90}\%$ \\ %
Inception-RN-v2~\cite{szegedy2017inception}\kern-.5em     & $75.59\%$ & $93.00\%$ & $\mathbf{75.73}\%$ & $\mathbf{93.32}\%$ \\ %
Xception~\cite{chollet2017xception}             & $77.57\%$ & $93.92\%$ & $\mathbf{77.71}\%$ & $\mathbf{93.97}\%$ \\ %
ResNet18~\cite{he2016deep}                      & $76.13\%$ & $93.01\%$ & $\mathbf{76.67}\%$ & $\mathbf{93.29}\%$ \\ %
ResNet34~\cite{he2016deep}                      & $77.34\%$ & $\mathbf{93.78}\%$ & $\mathbf{77.87}\%$ & ${93.75}\%$ \\ %
ResNet50~\cite{he2016deep}                      & $78.20\%$ & $94.28\%$ & $\mathbf{79.07}\%$ & $\mathbf{94.67}\%$ \\ %
ResNet101~\cite{he2016deep}                     & $79.07\%$ & $94.71\%$ & $\mathbf{79.51}\%$ & $\mathbf{94.87}\%$ \\ %
ResNet152~\cite{he2016deep}                     & $78.86\%$ & $94.65\%$ & $\mathbf{79.83}\%$ & $\mathbf{94.96}\%$ \\ %
ResNeXt50~\cite{xie2017aggregated}              & $78.55\%$ & $94.61\%$ & $\mathbf{78.92}\%$ & $\mathbf{94.80}\%$ \\ %
ResNeXt101~\cite{xie2017aggregated}             & $79.13\%$ & $\mathbf{94.85}\%$ & $\mathbf{79.54}\%$ & ${94.84}\%$ \\ %
ResNeXt152~\cite{xie2017aggregated}             & $79.26\%$ & $94.69\%$ & $\mathbf{79.48}\%$ & $\mathbf{94.89}\%$ \\ %
SE-ResNet18~\cite{hu2018squeeze}                & $76.25\%$ & $93.09\%$ & $\mathbf{76.77}\%$ & $\mathbf{93.36}\%$ \\ %
SE-ResNet34~\cite{hu2018squeeze}                & $77.85\%$ & $93.88\%$ & $\mathbf{78.20}\%$ & $\mathbf{94.13}\%$ \\ %
SE-ResNet50~\cite{hu2018squeeze}                & $77.78\%$ & $94.33\%$ & $\mathbf{78.79}\%$ & $\mathbf{94.53}\%$ \\ %
SE-ResNet101~\cite{hu2018squeeze}               & $77.94\%$ & $94.22\%$ & $\mathbf{79.19}\%$ & $\mathbf{94.70}\%$ \\ %
SE-ResNet152~\cite{hu2018squeeze}               & $78.10\%$ & $94.46\%$ & $\mathbf{79.35}\%$ & $\mathbf{94.73}\%$ \\ %
NASNet~\cite{zoph2018learning}                  & $77.76\%$ & $94.26\%$ & $\mathbf{78.17}\%$ & $\mathbf{94.35}\%$ \\ %
Wide-RN-40-10~\cite{zagoruyko2016wide}       & $78.93\%$ & $94.42\%$ & $\mathbf{79.60}\%$ & $\mathbf{94.80}\%$ \\ %
StochD-RN-18~\cite{huang2016deep}          & $75.39\%$ & $94.09\%$ & $\mathbf{75.44}\%$ & $\mathbf{94.13}\%$ \\ %
StochD-RN-34~\cite{huang2016deep}          & $78.03\%$ & $94.81\%$ & $\mathbf{78.16}\%$ & $\mathbf{94.97}\%$ \\ %
StochD-RN-50~\cite{huang2016deep}          & $77.02\%$ & $94.61\%$ & $\mathbf{77.40}\%$ & $\mathbf{94.78}\%$ \\ %
StochD-RN-101~\cite{huang2016deep}         & $78.72\%$ & $94.67\%$ & $\mathbf{78.96}\%$ & $\mathbf{94.75}\%$ \\ %
\midrule
Average                                    & $75.90\%$ & $93.19\%$ & $\mathbf{76.39}\%$ & $\mathbf{93.42}\%$ \\ %
\bottomrule
    \end{tabular}
    \vskip-0.2ex
    \caption{\label{tab:conv-cifar-100}%
    Results on \textbf{CIFAR-100} trained for $200$ epochs, averaged over 5 seeds. 
    The standard deviations and results for TrAct with only $133$ epochs are depicted in Tables~\ref{tab:conv-cifar-100-std} and~\ref{tab:conv-cifar-100-std-133} in the SM.
    }
    \vskip-3em
\end{wraptable}

ters as the reference, i.e., SGD with momentum ($0.9$), weight decay ($0.0005$), and learning rate schedule with $60$ epochs at $0.1$, $60$ epochs at $0.02$, $40$ epochs at $0.004$, $40$ epochs at $0.0008$, and a warmup schedule during the first epoch, for a total of $200$ epochs.
We reproduced each baseline on a set of 5 separate seeds, and discarded the models that produced NaNs on any of the 5 seeds of the baseline.
To make the evaluation feasible, we limit the hyperparameter for\parfillskip=0pt \par}

\vskip-\parskip
TrAct to $\lambda=0.1$.
Second, we also reproduce the ResNet-18 CIFAR-10 experiment but with CIFAR-100. The results for this are displayed in Figure~\ref{fig:resnet-18-cifar-100} in the Supplementary Material and demonstrate similar relations as the corresponding Figure~\ref{fig:resnet-18-cifar-10}.
Again, all models are trained with a batch size of $128$ on a single NVIDIA RTX 4090 GPU.

\paragraph{Results}
We display the results for the $36$ CIFAR-100 models in Table~\ref{tab:conv-cifar-100}.
We can observe that TrAct outperforms the baseline wrt.~top-1 and top-5 accuracy for $33$ and $34$ out of $36$ models, respectively.
Further, except for those $5$ models, for which TrAct and the baseline perform comparably (each better on one metric), TrAct is better than vanilla training.
Specifically, for $31$ models, TrAct outperforms the baseline on both metrics, and the overall best result is also achieved by TrAct.
Further, TrAct improves the accuracy on average by $0.49\%$ on top-1 accuracy and by $0.23\%$ on top-5 accuracy, a statistically very significant improvement over the baseline. The average standard deviations are $0.25\%$ and $0.15\%$ for top-1 and top-5 accuracy, respectively.

In addition, we also considered training the models with TrAct for only $133$ epochs, i.e., $2/3$ of the training time.
Here, we found that, on average, regular training for $200$ epochs is comparable with TrAct for $133$ epochs with a small advantage for TrAct. 
In particular, the average accuracy of TrAct with $133$ epochs is $75.94\%$ (top-1) and $93.34\%$ (top-5), which is a small improvement over regular training for $200$ epochs. %
The individual results are reported in Table~\ref{tab:conv-cifar-100-std-133} in the Supplementary Material.

\vspace{-.4em}
\subsection{ImageNet}
\vspace{-.2em}

Finally, we consider training on the ImageNet data set~\cite{deng2009imagenet}. 
We train ResNet-$\{18, 34, 50\}$, ViT-S and ViT-B models.

\vspace{-.5em}
\paragraph{ResNet Setup}
For the ResNet-$\{18, 34, 50\}$ models, we train for $\{30, 60, 90\}$ epochs and consider base learning rates in the grid $\{0.2, 0.141, 0.1, 0.071, 0.05\}$ and determine the choice for each model / training length combination with standard baseline training. 
We find that for each model, when training for $30$ epochs, $0.141$ performs best, and, when training for $\{60, 90\}$ epochs, $0.1$ performs best as the base learning rate.
We use SGD with momentum ($0.9$), weight decay ($0.0001$), and the typical learning rate schedule, which decays the learning rate after $1/3$ and $2/3$ of training by $0.1$ each.
For TrAct, we (again) use the same learning rate as optimal for the baseline, and consider $\lambda\in\{0.05, 0.1, 0.2\}$.
Each ResNet model is trained with a batch size of $256$ on a single NVIDIA RTX 4090 GPU.

\begin{figure}[h]%
    \centering
    \includegraphics[width=.49\linewidth]{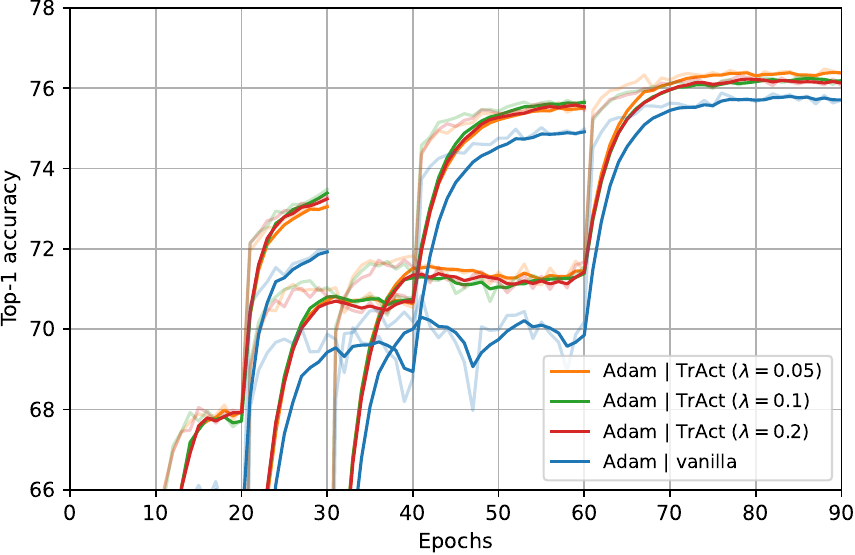}\hfill
    \includegraphics[width=.49\linewidth]{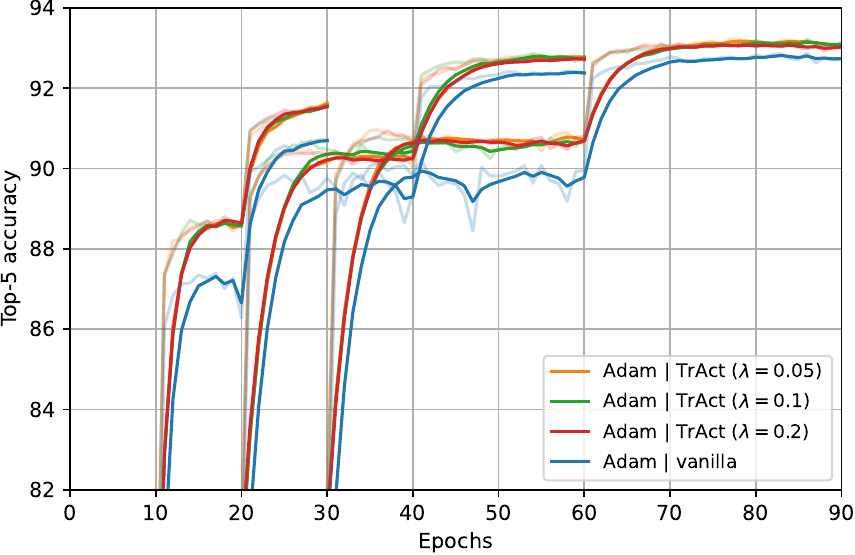}
    \vspace{-.5em}
    \captionof{figure}{
    Test accuracy of \textbf{ResNet-50} trained on \textbf{ImageNet} for $\{30, 60, 90\}$ epochs.
    When training for $60$ epochs with TrAct, we achieve comparable accuracy to standard training for $90$ epochs, showing a $1.5\times$ speedup. 
    Plots for ResNet-18/34 are in the SM.
    \label{fig:resnet-50-imagenet}
    }
    \vspace{-.25em}
\end{figure}

\begin{wraptable}[9]{r}{0.53\textwidth}
    \centering
    \vspace{-.75em}
    \setlength{\tabcolsep}{4.pt}
    \footnotesize
    \begin{tabular}{c|cc|cc}
\toprule
                                              & \multicolumn{2}{c}{Baseline} & \multicolumn{2}{c}{TrAct ($\lambda{=}0.1$)} \\
Num. epochs                                           & Top-1 & Top-5 & Top-1 & Top-5 \\
\midrule
30         & $71.96\%$ & $90.70\%$ & $\mathbf{73.48}\%$ & $\mathbf{91.61}\%$ \\
60         & $74.98\%$ & $92.36\%$ & \cellcolor{almond}$\mathbf{75.68}\%$ & \cellcolor{almond}$\mathbf{92.78}\%$ \\
90         & \cellcolor{almond}$75.70\%$ & \cellcolor{almond}$92.74\%$ & $\mathbf{76.20}\%$ & $\mathbf{93.12}\%$ \\
\bottomrule
    \end{tabular}
    \vspace{-0.3em}
    \captionof{table}{\label{tab:resnet-50-imagenet}%
    Final test accuracies (ImageNet valid~set) for training \textbf{ResNet-50}~\cite{he2016deep} on \textbf{ImageNet}. 
    TrAct with only 60 epochs performs comparable to the baseline with 90 epochs. 
    }
    \vskip-1em
\end{wraptable}
\vspace{-.5em}
\paragraph{ResNet Results}
We start by discussing the ResNet results and then proceed with the vision transformers.
We present training plots for%
ResNet-50 in Figure~\ref{fig:resnet-50-imagenet}.
Here, we can observe an effective speedup of a factor of $1.5$ during training, which we also demonstrate in Table~\ref{tab:resnet-50-imagenet}.
In particular, the difference in accuracy for TrAct ($\lambda=0.1$) with $60$ compared to the baseline with full $90$ epoch training is $-0.02\%$ and $+0.04\%$ for top-1 and top-5.

\paragraph{ViT Setup}
For training the ViTs, we reproduce the ``DeiT III''~\cite{Touvron_et_al_2023}, which provides the strongest baseline that is reproducible on a single 8-GPU node.
We train each model with the same hyperparameters as in the official source code\footnote{Based on \href{https://github.com/facebookresearch/deit}{github.com/facebookresearch/deit}.}.
We note that the ViT-S and ViT-B are both trained at a batch size of $2\,048$ and are pre-trained on resolutions of $224$ and $192$, respectively, and both models are finetuned on a resolution of $224$.
We consider pre-training for $400$ and $800$ epochs.
Finetuning for each model is performed for $50$ epochs.
For the $400$ epoch pre-training with TrAct, we use the stronger $\lambda=0.1$, while for the longer $800$ epoch pre-training we use the weaker $\lambda=0.2$.
We train the ViT-S models on $4$ NVIDIA A40 GPUs and the ViT-B models on $8$ NVIDIA V100 (32GB) GPUs.

\begin{wraptable}[15]{r}{0.55\textwidth}
    \centering
    \vspace{-.5em}
    \setlength{\tabcolsep}{5pt}
    {\small
    \begin{tabular}{llcccccc}
    \toprule
        DeiT-III Model                      & Epochs & Top-1 & Top-5  \\
    \midrule
        ViT-S~\cite{Touvron_et_al_2023}     & 400ep & $80.4\%$ & ---  \\ %
        ViT-S~$\dagger$                     & 400ep & $81.23\%$ & $95.70\%$ \\ %
        \rowcolor{almond} ViT-S + TrAct ($\lambda{=}0.1$)     & 400ep & $\mathbf{81.50}\%$ & $\mathbf{95.73}\%$ \\ %
    \midrule
        ViT-S~\cite{Touvron_et_al_2023}     & 800ep & $81.4\%$ & ---  \\ %
        ViT-S~$\dagger$                     & 800ep & $81.97\%$ & $95.90\%$  \\ %
        \rowcolor{almond} ViT-S + TrAct ($\lambda{=}0.2$)     & 800ep & $\mathbf{82.18}\%$ & $\mathbf{95.98}\%$  \\ %
    \midrule
        ViT-B~\cite{Touvron_et_al_2023}     & 400ep & $83.5\%$ & ---  \\ %
        ViT-B~$\dagger$                     & 400ep & $83.34\%$ & $96.44\%$ \\ %
        \rowcolor{almond} ViT-B + TrAct ($\lambda{=}0.1$)     & 400ep & $\mathbf{83.58}\%$ & $\mathbf{96.52}\%$ \\ %
    \bottomrule
    \end{tabular}}
    \captionof{table}{\label{tab:vit-imagenet}%
    Results for training \textbf{ViTs} (DeiT-III) on \textbf{ImageNet-1k}. 
    $\dagger$ denotes our reproduction.
    }
\end{wraptable}

\vspace{-.5em}
\paragraph{ViT Results}
In Table~\ref{tab:vit-imagenet} we present the results for training vision transformers.
First, we observe that our reproductions following the official code and hyperparameters improved over the originally reported baselines, potentially due to contemporary improvements in the underlying libraries (our hardware only supported more recent versions).
Notably, TrAct consistently improves upon our improved baselines.
{We note that we did not change any hyperparameters for training with TrAct.}
For ViT-S, using TrAct leads to $36\%$ of the improvement that can be achieved by training the 
baseline twice as long.
These improvements can be considered quite substantial considering that these are very large models and we modified only the training of the first layer.
Notably, here, the runtime overheads were particularly small, ranging from $0.08\%$ to $0.25\%$.
{Finally, we consider the quality of the pre-trained model outside\parfillskip=0pt \par}

\begin{wraptable}[7]{r}{0.55\textwidth}
    \setlength{\tabcolsep}{2pt}
    {\small
    \begin{tabular}{lccccccc}
    \toprule
        Model/Dataset                      & CIFAR-10  & CIFAR-100 & Flowers & S.~Cars \\
    \midrule
        ViT-S                               & $98.94\%$ & $90.70\%$ & $94.39\%$ & $90.44\%$ \\ %
        \rowcolor{almond} ViT-S + TrAct     & $\mathbf{99.02}\%$ & $\mathbf{90.85}\%$ & $\mathbf{95.58}\%$ & $\mathbf{91.07}\%$ \\ %
    \bottomrule
    \end{tabular}}
    \captionof{table}{\label{tab:vit-transfer}%
    Transfer learning results for ViT-S on CIFAR-10 and CIFAR-100~\cite{krizhevsky2009cifar10}, Flowers-102~\cite{nilsback2008automated}, and Stanford Cars~\cite{krause2013collecting}.
    }
\end{wraptable}

of ImageNet. 
We fine-tune the ViT-S ($800$ epoch pre-training) model on the data sets CIFAR-10 and CIFAR-100~\cite{krizhevsky2009cifar10} ($200$ epochs), Flowers-102~\cite{nilsback2008automated} ($5000$ epochs), and Stanford Cars~\cite{krause2013collecting} ($1000$ epochs).
For the baseline, both pre-training and fine-tuning were performed with the vanilla method, and, for TrAct, both pre-training and fine-tuning were performed with TrAct.
In Table~\ref{tab:vit-transfer}, we can observe consistent improvements for training with TrAct.

\begin{wrapfigure}[10]{r}{0.5\textwidth}
    \centering
    \vspace{-1.35em}
    \includegraphics[width=\linewidth]{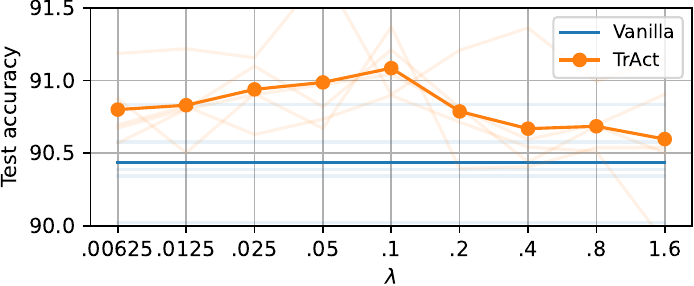}
    \vspace{-1.5em}
    \caption{\label{fig:vit-cifar-10-lambda}%
    Effect of ${\lambda}$ for training a ViT on CIFAR-10. Training for $200$ ep., setup as Fig.~\ref{fig:vit-cifar-10}, avg.\ over 5 seeds.
    }
    \vskip-1em
\end{wrapfigure}
\vspace{-.5em}
\subsection{Effect of $\pmb{\lambda}$}
\vspace{-.3em}
$\lambda$ is the only hyperparameter introduced by TrAct.
Often, with an additional hyperparameter, the hyperparameter space becomes more difficult to manage.
However, for TrAct, the selection of $\lambda$ is simple and compatible with existing hyperparameters.
Therefore, throughout all experiments in this paper, we kept all other hyperparameters equal to the optimal choice for the respective baselines, and only considered $\lambda\in\{0.05, 0.1, 0.2\}$.
A general trend is that with smaller $\lambda$s, TrAct becomes more aggressive, which tends to be more favorable in shorter training, and for larger $\lambda$s, TrAct is more moderate, which is ideal for longer trainings.
However, in many cases, the particular choice of $\lambda\in\{0.05, 0.1, 0.2\}$ has only a subtle impact on accuracy as can be seen throughout the figures in this work. 
Further, going beyond this range of $\lambda$s, in Fig.~\ref{fig:vit-cifar-10-lambda}, we can observe that TrAct is robust against changes in this parameter.
In all experiments, the data was as-per-convention standardized to mean $0$ and standard deviation $1$; deviating from this convention could change the space of $\lambda$s.
For significantly different tasks and drastically different kernel sizes or number of input channels, we expect that the space of $\lambda$s could change.
Overall, we recommend $\lambda=0.1$ as a starting point and, for long training, we recommend $\lambda=0.2$.

\vspace{-.5em}
\subsection{Ablation Study}
\vspace{-.3em}

As an ablation study, we first compare TrAct to patch-wise layer normalization for ViTs.
For this, we normalize the pixel values of each input patch to mean~$0$ and standard deviation~$1$.
This is an alternate solution to the conceptual problem of low contrast image regions having a lesser effect on the first layer optimization compared to higher contrast image regions.
However, here, we also note that, in contrast to TrAct, the actual neural network inputs are changed through the normalization.
Further, we consider DualPatchNorm~\cite{Kumar_et_al_2023} as a comparison, which additionally includes a second patch normalization layer after the first linear layer, and introduces additional trainable weight parameters for affine transformations into both patch normalization layers.

\begin{wrapfigure}[15]{r}{0.5\textwidth}
    \centering
    \vspace{-1em}
    \includegraphics[width=\linewidth]{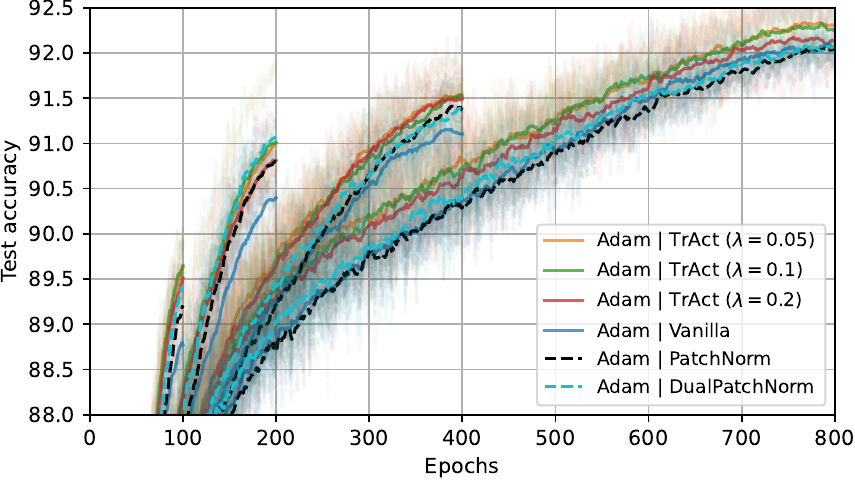}
    \vskip-0.7ex
    \caption{\label{fig:vit-cifar-10-whitening}Ablation Study: training a ViT on CIFAR-10, including patch normalization (black, dashed) and DualPatchNorm (cyan, dashed). 
    Setups as in Figure~\ref{fig:vit-cifar-10}, averaged over $5$ seeds.
    }
    \vskip-1em
\end{wrapfigure}
We use the same setup as for the CIFAR-10 ViT and run each setting for $5$ seeds.
The results are displayed in Figure~\ref{fig:vit-cifar-10-whitening}.
Here, we observe that patch normalization improves training for up to $400$ epochs compared to the baseline; however, not as much as TrAct does. 
{Further, we find that} %
DualPatchNorm performs equivalently compared to input patch normalization and worse than TrAct, except for the case of $200$ epochs where it performs insignificantly better than TrAct. 
For training for $800$ epochs, patch normalization and DualPatchNorm do not improve the baseline and perform insignificantly worse, whereas TrAct still shows accuracy improvements.
This effect may be explained by the fact that patch normalization is a scalar form of whitening, and whitening can hurt generalization capabilities due to a loss of information~\cite{wadia2021whitening}.
In particular, what may be problematic is that patch normalization also affects the model behavior during inference, which contrasts TrAct.

As a second ablation study, we examine what happens if we (against convention) do not perform standardization of the data set.
We train the same ViTs as above on CIFAR-10 for $200$ epochs, averaged over $5$ seeds.
We consider two cases: first, an input value range of $[0,1]$ and a quite extreme input value range of $[0,255]$.

\begin{wrapfigure}[13]{r}{0.5\textwidth}
    \centering
    \vskip-2.75ex
    \includegraphics[width=\linewidth]{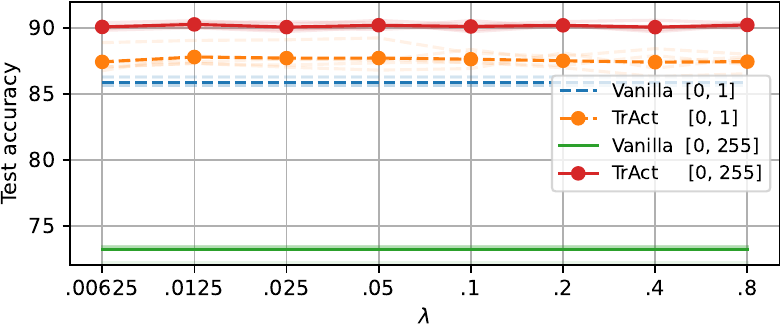}
    \vspace{-1.5em}
    \caption{\label{fig:vit-cifar-10-input-ranges}Ablation Study: training a ViT on CIFAR-10 \textit{without} data standardization and with input value ranges of $[0,1]$ vs.\ $[0,255]$. 
    Setups as in Figure~\ref{fig:vit-cifar-10}, $200$ epochs, and avg.\ over $5$ seeds. All other experiments in this work are trained \textit{with} data standardization.
    }
\end{wrapfigure}

We display the results in Figure~\ref{fig:vit-cifar-10-input-ranges}.
Here, we observe that TrAct is more robust against a lack of standardization.
Interestingly, we observe that TrAct performs better for the range of $[0,255]$ than $[0,1]$. 
The reason for this is that TrAct suffers from obtaining only positive inputs, which affects the $xx^\top$ matrix in Equation~\ref{eq:w-update-full}; however, we note that regular training suffers even more from the lack of standardization.
When considering the range of $[0,255]$, we observe that TrAct is virtually agnostic to $\lambda$, which is caused by the $xx^\top\!$ matrix becoming very large.
The reason why TrAct performs so well here (compared to the baseline) is that, due to the large $xx^\top\!$, the updates $\Delta W$ become very small. 
This is more desirable compared to the standard gradient, which explodes due to its proportionality to the input values, and therefore drastically degrades training.

\begin{wrapfigure}[13]{r}{0.5\textwidth}
    \centering
    \vspace{-1.5em}
    \includegraphics[width=\linewidth]{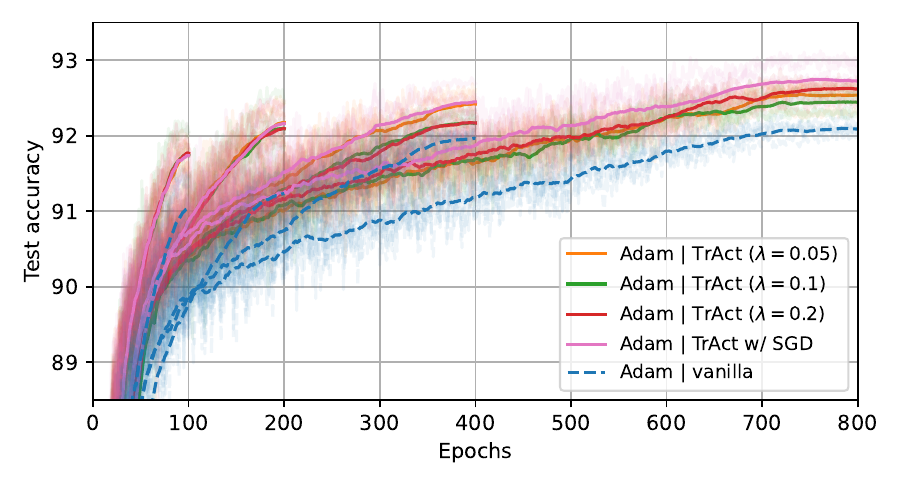}
    \vspace{-1.75em}
    \caption{
    Ablation Study: extending Figure \ref{fig:resnet-18-cifar-10} (right) by training the first layer with TrAct and SGD (pink) and the remainder of the model still with Adam. 
    \label{fig:sgdadam}
    }
\end{wrapfigure}
In each experiment, we used only a single optimizer for the entire model; however, our theory assumes that TrAct is used with SGD.
This motivates the question of whether it is advantageous to train the first layer with SGD, while training the remainder of the model, e.g., with Adam.

Thus, as a final ablation study, we extend the experiment from Figure~\ref{fig:resnet-18-cifar-10} (right) by training the first layer with SGD while training the remaining model with Adam. 
We display the results in Figure~\ref{fig:sgdadam} where we can observe small improvements when using SGD for the TrAct layer.

\subsection{Runtime Analysis}

In this section, we provide a training runtime analysis. 
Overall, the trend is that, for large models, TrAct adds on a tiny runtime overhead, while it can become more expensive for smaller models. 
In particular, for the CIFAR-10 ViT, the average training time per $100$ epochs increased by $9.7\%$ from $1091$s to $1197$s.
Much of this can be attributed to the required additional CUDA calls and non-fused operations, which can be expensive for cheaper tasks.
However, when considering larger models, this overhead almost entirely amortizes.
In particular the ViT-S ($800$ epochs) pre-training cost increased by only $0.08\%$ from 133:52 hours to 133:58 hours.
The pre-training cost of the ViT-B ($400$ epochs) increased by $0.25\%$ from 98:28 hours to 98:43 hours.
We can see that, in each case, the training cost overhead is clearly more than worth the reduced requirement of epochs already.
Further, fused kernels could drastically reduce the computational overhead; in particular, our current implementation replaces an existing fused operation by multiple calls from the Python space.
As TrAct only affects training, and the modification isn't present during forwarding, TrAct has no effect on inference time.

\section{Discussion \& Conclusion}
\label{sec:conclusion}
In this work, we introduced TrAct, a novel training strategy that modifies the optimization behavior of the first layer, leading to significant performance improvements across a range of $50$ experimental setups.
The approach is efficient and effectively speeds up training by factors between $1.25\times$ and $4\times$ depending on the model size.
We hope that the simplicity of integration into existing training schemes as well as the robust performance improvements motivate the community to adopt TrAct.

\begin{ack}
This work was in part supported by the Federal Agency for Disruptive Innovation SPRIN-D, the Land Salzburg within the WISS 2025 project IDA-Lab (20102-F1901166-KZP and 20204-WISS/225/197-2019), the ARO~(W911NF-21-1-0125), the ONR~(N00014-23-1-2159), and the CZ~Biohub.
\end{ack}

\printbibliography

\clearpage
\appendix

\section{Theory}
\label{sm:theory}
\setcounter{lemma}{0}

\begin{lemma}
The solution $\Delta W_i^\star$ of Equation~\ref{eq:problem-wi} is
\begin{align}
    \Delta W_i^\star
    &= -\,\eta \cdot \nabla_{z_i} \mathcal{L}(z)  \cdot x^\top \cdot \left(\frac{xx^\top\!}{b} + \lambda\cdot I_n\right)^{-1} \!.
\end{align}
\begin{proof}

We would like to solve the optimization problem
    \[ \arg\min_{\Delta W_i} ~
   \| - \eta b\, \nabla_{\!z_i} \mathcal{L} (z)
   - \Delta W_{\!i} x \|_2^2
   \,+\, \lambda b\, \| \Delta W_{\!i} \|_2^2. \]
A necessary condition for a minimum of the functional
\[ F(\Delta W_{\!i}) = 
   (-\eta b\, \nabla_{\!z_i} \mathcal{L}(z) -\Delta W_{\!i} x)^2
   \,+\, \lambda b \, (\Delta W_{\!i}) (\Delta W_{\!i})^\top \]
is that $\nabla_{\!\Delta W_i} F(\Delta W_i)$ vanishes:
\begin{eqnarray*}
\nabla_{\!\Delta W_i} F(\Delta W_i)
& = & \nabla_{\!\Delta W_i} 
      (-\eta b\,\nabla_{\!z_i} \mathcal{L}(z) -\Delta W_{\!i} x)^2
      +\lambda b \nabla_{\!\Delta W_i}
      ((\Delta W_i)(\Delta W_{\!i})^\top) \\
& = & 2 (-\eta b\,\nabla_{\!z_i} \mathcal{L}(z) -\Delta W_{\!i} x)
      (\nabla_{\!\Delta W_i}
      (-\eta b\, \nabla_{\!z_i} \mathcal{L}(z) -\Delta W_{\!i} x))
      + 2\lambda b\, \Delta W_i \\
& = & 2 (-\eta b\, \nabla_{\!z_i} \mathcal{L}(z) -\Delta W_{\!i} x)
      (-x)^\top
  +   2\lambda b\, \Delta W_{\!i} \\
& = & 2 (\eta b\, \nabla_{\!z_i} \mathcal{L}(z) + \Delta W_{\!i} x)
      \, x^\top
  +   2\lambda b\, \Delta W_{\!i}
      \quad\stackrel{!}{=}\quad 0.
\end{eqnarray*}
It follows for the optimal $\Delta W_{\!i}^\star$ that minimizes
$F(\Delta W_{\!i})$
\begin{eqnarray*}
\lefteqn{\eta b\, (\nabla_{\!z_i} \mathcal{L}(z)) \, x^\top
         + \Delta W_{\!i}^\star x x^\top
         + \lambda b\, \Delta W_{\!i}^\star = 0} \\
& \Leftrightarrow &
-\eta b\, (\nabla_{\!z_i} \mathcal{L}(z)) \, x^\top
  = \Delta W_{\!i}^\star (x x^\top + \lambda b I_n) \\
& \Leftrightarrow &
  \Delta W_{\!i}^\star = 
  -\eta b\, (\nabla_{\!z_i} \mathcal{L}(z)) \, x^\top
  (x x^\top + \lambda b I_n)^{-1} \\
& \Leftrightarrow &
  \Delta W_{\!i}^\star = 
  -\eta\, (\nabla_{\!z_i} \mathcal{L}(z)) \, x^\top
  \Big(\frac{x x^\top}{b} + \lambda I_n\Big)^{\!-1}.
\end{eqnarray*}
\end{proof}
\end{lemma}

\begin{lemma}
    Using TrAct does not change the set of possible convergence points compared to vanilla (full batch) gradient descent.
    Herein, we use the standard definition of convergence points as those points where no update is performed because the gradient is zero.
    \begin{proof}
    First, we remark that only the training of the first layer is affected by TrAct.
    To show the statement, we show that $(i)$ a zero gradient for GD implies that TrAct also performs no update and that $(ii)$ TrAct performing no update implies zero gradients for GD.
    \\

    $(i)$~~In the first case, we assume that gradient descent has converged, i.e., the gradient wrt.~first layer weights is zero $\nabla_{W} \mathcal{L}(W)=\mathbf{0}$.
    We want to show that, in this case, our proposed update is also zero, i.e., $\Delta W^\star=\mathbf{0}$.
    Using the definition of $\Delta W^\star$ from Equation~\ref{eq:combining-rows-from-lemma-1}, we have
        \begin{align}
            \Delta W^\star 
            &= -\,\eta  \cdot \nabla_{z} \mathcal{L}(z)  \cdot x^\top \cdot \left(\frac{xx^\top\!}{b} + \lambda\cdot I_n\right)^{-1}\\
            &= -\,\eta  \cdot \nabla_{W} \mathcal{L}(W) \cdot \left(\frac{xx^\top\!}{b} + \lambda\cdot I_n\right)^{-1}\\
            &= -\,\eta  \cdot \mathbf{0} \cdot \left(\frac{xx^\top\!}{b} + \lambda\cdot I_n\right)^{-1} = \mathbf{0}\,,
        \end{align}
    which shows this direction.
    \\

    $(ii)$~~In the second case, we have $\Delta W^\star=\mathbf{0}$ and need to show that this implies $\nabla_{W} \mathcal{L}(W)=\mathbf{0}$.
    For this, we can observe that $\left(xx^\top /\, b + \lambda\cdot I_n\right)^{-1}$ is PD (positive definite) by definition and $\left(xx^\top /\, b + \lambda\cdot I_n\right)$ also exists.
    If $\Delta W^\star=\mathbf{0}$, then
    \begin{align}
        \mathbf{0} 
        &= \Delta W^\star= \Delta W^\star \left(\frac{xx^\top\!}{b} + \lambda\cdot I_n\right) \\
        &= -\eta {\cdot} \nabla_{z} \mathcal{L}(z)  {\cdot} x^\top \!\!\cdot\! \left(\frac{xx^\top\!}{b} + \lambda\cdot I_n\right)^{\!\!-1}\!\!\!\! \left(\frac{xx^\top\!}{b} + \lambda\cdot I_n\right)\notag\\
        &= -\,\eta  \cdot \nabla_{z} \mathcal{L}(z)  \cdot x^\top =  -\,\eta  \cdot \nabla_{W} \mathcal{L}(W)\,,
    \end{align}
    which also shows this direction.
    \\

    Overall, we showed that if gradient descent has converged according to the standard notion of a zero gradient, then our update has also converged and vice versa.
    
    \end{proof}
\end{lemma}

\section{Additional Results}

\begin{figure*}
    \centering
    \includegraphics[width=.49\linewidth]{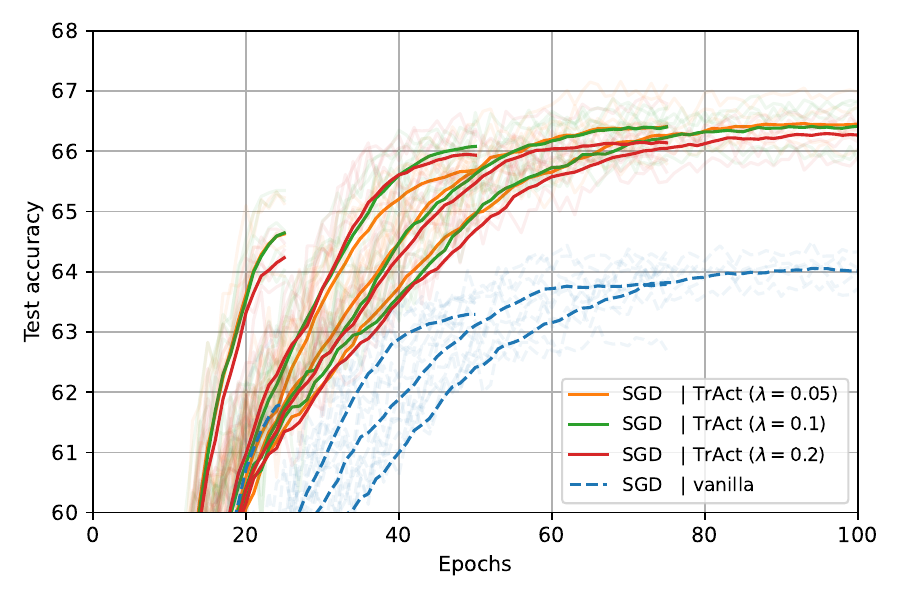}\hfill
    \includegraphics[width=.49\linewidth]{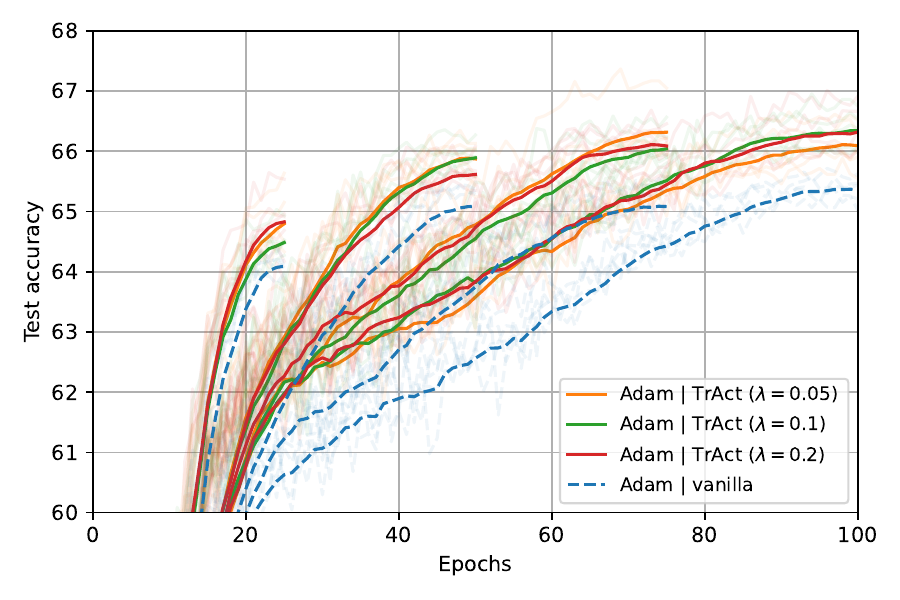}
    \vspace{-1em}
    \caption{
    Training a \textbf{ResNet-18} on \textbf{CIFAR-100} with the CIFAR-10 setup from Section~\ref{sec:cifar-10}. Displayed is top-1 accuracy. 
    We train for $\{100,200,400,800\}$ epochs using a cosine learning rate schedule and with SGD (left) and Adam (right). 
    Learning rates have been selected as optimal for each baseline. 
    Averaged over 5 seeds. 
    TrAct (solid lines) consistently outperforms the baselines (dashed lines).
    }
    \label{fig:resnet-18-cifar-100}
\end{figure*}

\begin{figure*}[t!]
    \centering
    \includegraphics[width=.49\linewidth]{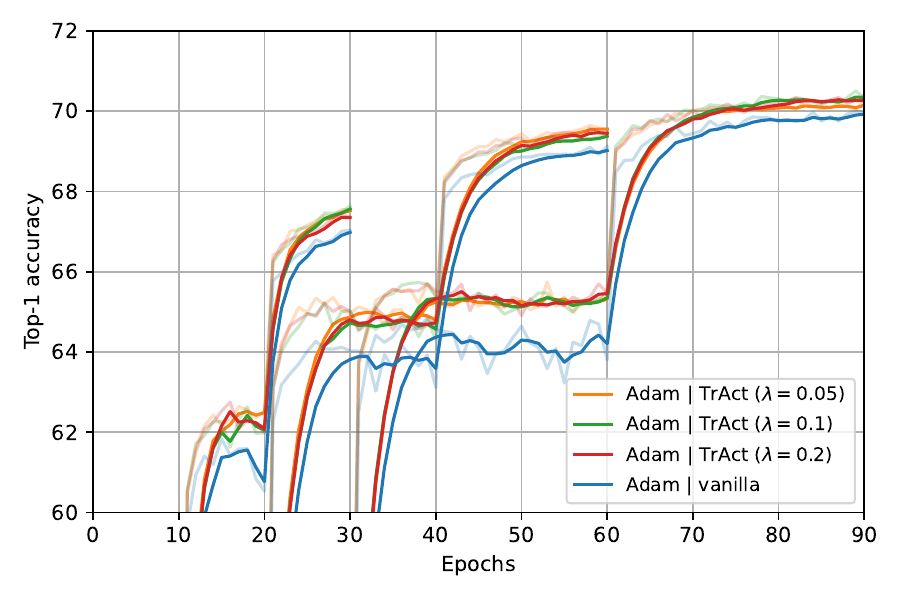}\hfill%
    \includegraphics[width=.49\linewidth]{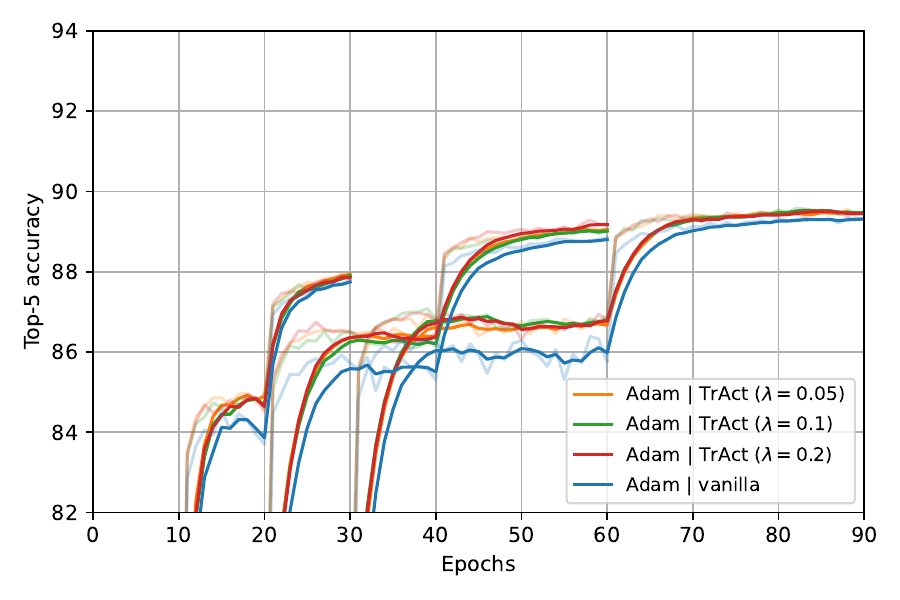}
    \vspace{-1em}
    \caption{Test accuracy of \textbf{ResNet-18} trained on \textbf{ImageNet} for $\{30, 60, 90\}$ epochs. Displayed is the top-1 (left) and top-5 (right) accuracy.}
    \label{fig:resnet-18-imagenet}
\end{figure*}

\begin{figure*}[t!]
    \centering
    \includegraphics[width=.49\linewidth]{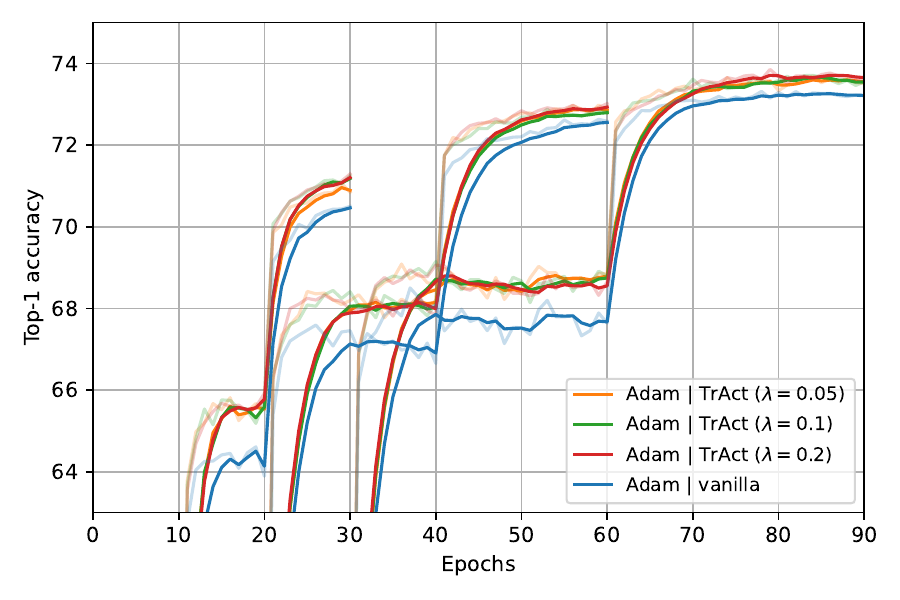}\hfill%
    \includegraphics[width=.49\linewidth]{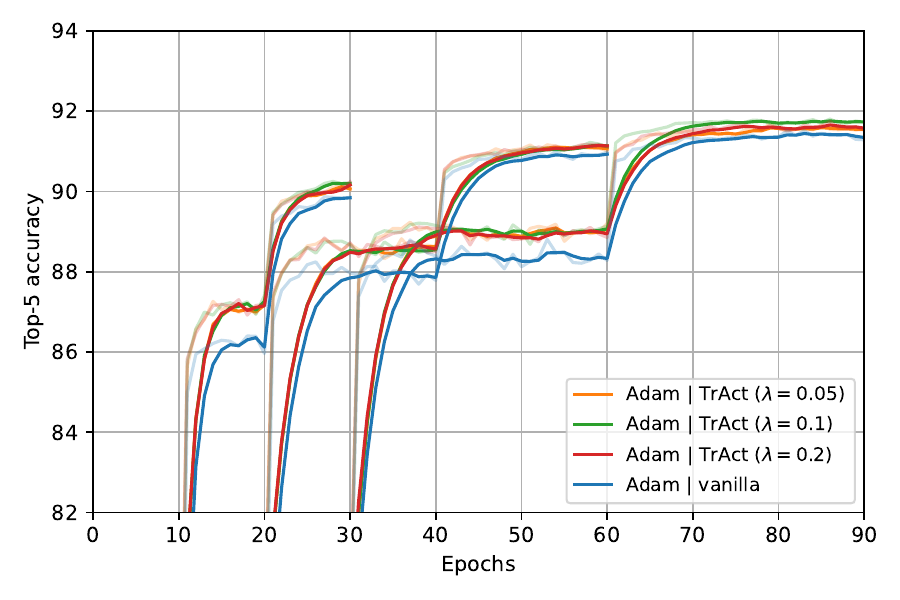}
    \vspace{-1em}
    \caption{Test accuracy of \textbf{ResNet-34} trained on \textbf{ImageNet} for $\{30, 60, 90\}$ epochs. Displayed is the top-1 (left) and top-5 (right) accuracy.}
    \label{fig:resnet-34-imagenet}
\end{figure*}

We display additional results in Figures~\ref{fig:resnet-18-cifar-100}, \ref{fig:resnet-18-imagenet}, and~\ref{fig:resnet-34-imagenet} as well as in Tables~\ref{tab:conv-cifar-100-std} and~\ref{tab:conv-cifar-100-std-133}.
\\

As an additional experiment, in order to verify the applicability of TrAct beyond training / pre-training, we train Faster R-CNN models \cite{Ren_et_al_2015} on PASCAL VOC2007 \cite{pascal-voc-2007} using a VGG-16 backbone~\cite{simonyan2014very}. 
However, Faster R-CNN uses a pretrained vision encoder where the first 4 layers are frozen. 
In order to enable TrAct, as TrAct only affects the training of the first layer, we unfreeze these first layers when training the object detection head. 
The mean average precision (mAP) on test data for the vanilla model versus TrAct training are shown in Table~\ref{tab:map_fasterrcnn}.

\begin{table}[ht]
\centering
\begin{minipage}{.5\linewidth}
    \centering
    \begin{tabular}{cc}
    \toprule
    vanilla & TrAct \\
    \midrule
    $0.659 \pm 0.005$ & $0.671 \pm 0.004$ \\
    \bottomrule
    \end{tabular}
    \vspace{1em}
    \caption{\label{tab:map_fasterrcnn}Mean average precision (mAP) on test data for Faster R-CNN \cite{Ren_et_al_2015} with a VGG-16 backbone on PASCAL VOC2007 \cite{pascal-voc-2007}, averaged over 2 seeds.}
    \vspace{-1em}
\end{minipage}
\end{table}

We can observe that TrAct performs better than the vanilla method by about 1.1\%.
We would like to point out that, while is TrAct especially designed for speeding up pretraining or training from scratch, i.e., when actually learning the first layer, we find that it also helps in finetuning pretrained models. 
Here, a limitation is of course that TrAct requires actually training the first layer.

\clearpage

\begin{table*}[]
    \centering
    \setlength{\tabcolsep}{6pt}
    \small
    \begin{tabular}{l|cc|cc}
\toprule
                                                & \multicolumn{2}{c}{Baseline} & \multicolumn{2}{c}{TrAct ($\lambda{=}0.1$)} \\
Model                                           & Top-1 & Top-5 & Top-1 & Top-5 \\
\midrule
SqueezeNet~\cite{iandola2016squeezenet}         & $69.45\% \pm 0.30\%$ & $91.09\% \pm 0.20\%$ & $70.48\% \pm 0.17\%$ & $91.50\% \pm 0.13\%$ \\ %
MobileNet~\cite{howard2017mobilenets}           & $66.99\% \pm 0.16\%$ & $88.95\% \pm 0.07\%$ & $67.06\% \pm 0.41\%$ & $89.12\% \pm 0.16\%$ \\ %
MobileNetV2~\cite{sandler2018mobilenetv2}       & $67.76\% \pm 0.20\%$ & $90.80\% \pm 0.10\%$ & $67.89\% \pm 0.22\%$ & $90.91\% \pm 0.11\%$ \\ %
ShuffleNet~\cite{zhang2018shufflenet}           & $69.98\% \pm 0.22\%$ & $91.18\% \pm 0.12\%$ & $69.97\% \pm 0.30\%$ & $91.45\% \pm 0.29\%$ \\ %
ShuffleNetV2~\cite{ma2018shufflenet}            & $69.31\% \pm 0.13\%$ & $90.91\% \pm 0.15\%$ & $69.88\% \pm 0.26\%$ & $91.02\% \pm 0.08\%$ \\ %
VGG-11~\cite{simonyan2014very}                  & $68.44\% \pm 0.24\%$ & $88.02\% \pm 0.10\%$ & $69.66\% \pm 0.20\%$ & $88.99\% \pm 0.21\%$ \\ %
VGG-13~\cite{simonyan2014very}                  & $71.96\% \pm 0.26\%$ & $90.27\% \pm 0.17\%$ & $72.98\% \pm 0.18\%$ & $90.78\% \pm 0.15\%$ \\ %
VGG-16~\cite{simonyan2014very}                  & $72.12\% \pm 0.24\%$ & $89.81\% \pm 0.19\%$ & $72.73\% \pm 0.16\%$ & $90.11\% \pm 0.15\%$ \\ %
VGG-19~\cite{simonyan2014very}                  & $71.13\% \pm 0.46\%$ & $88.10\% \pm 0.36\%$ & $71.45\% \pm 0.34\%$ & $88.42\% \pm 0.46\%$ \\ %
DenseNet121~\cite{huang2017densely}             & $78.93\% \pm 0.28\%$ & $94.83\% \pm 0.13\%$ & $79.55\% \pm 0.25\%$ & $94.92\% \pm 0.11\%$ \\ %
DenseNet161~\cite{huang2017densely}             & $79.95\% \pm 0.21\%$ & $95.25\% \pm 0.19\%$ & $80.47\% \pm 0.25\%$ & $95.37\% \pm 0.12\%$ \\ %
DenseNet201~\cite{huang2017densely}             & $79.39\% \pm 0.20\%$ & $95.07\% \pm 0.12\%$ & $79.94\% \pm 0.19\%$ & $95.17\% \pm 0.10\%$ \\ %
GoogLeNet~\cite{szegedy2014going}               & $76.85\% \pm 0.14\%$ & $93.53\% \pm 0.16\%$ & $77.18\% \pm 0.11\%$ & $93.86\% \pm 0.10\%$ \\ %
Inception-v3~\cite{szegedy2016rethinking}       & $79.40\% \pm 0.15\%$ & $94.94\% \pm 0.21\%$ & $79.24\% \pm 0.33\%$ & $95.04\% \pm 0.06\%$ \\ %
Inception-v4~\cite{szegedy2017inception}        & $77.32\% \pm 0.36\%$ & $93.80\% \pm 0.33\%$ & $77.14\% \pm 0.28\%$ & $93.90\% \pm 0.20\%$ \\ %
Inception-RN-v2~\cite{szegedy2017inception}     & $75.59\% \pm 0.45\%$ & $93.00\% \pm 0.18\%$ & $75.73\% \pm 0.30\%$ & $93.32\% \pm 0.19\%$ \\ %
Xception~\cite{chollet2017xception}             & $77.57\% \pm 0.31\%$ & $93.92\% \pm 0.17\%$ & $77.71\% \pm 0.17\%$ & $93.97\% \pm 0.10\%$ \\ %
ResNet18~\cite{he2016deep}                      & $76.13\% \pm 0.27\%$ & $93.01\% \pm 0.06\%$ & $76.67\% \pm 0.26\%$ & $93.29\% \pm 0.22\%$ \\ %
ResNet34~\cite{he2016deep}                      & $77.34\% \pm 0.33\%$ & $93.78\% \pm 0.16\%$ & $77.87\% \pm 0.25\%$ & $93.75\% \pm 0.10\%$ \\ %
ResNet50~\cite{he2016deep}                      & $78.20\% \pm 0.35\%$ & $94.28\% \pm 0.09\%$ & $79.07\% \pm 0.18\%$ & $94.67\% \pm 0.07\%$ \\ %
ResNet101~\cite{he2016deep}                     & $79.07\% \pm 0.22\%$ & $94.71\% \pm 0.20\%$ & $79.51\% \pm 0.43\%$ & $94.87\% \pm 0.06\%$ \\ %
ResNet152~\cite{he2016deep}                     & $78.86\% \pm 0.28\%$ & $94.65\% \pm 0.22\%$ & $79.83\% \pm 0.22\%$ & $94.96\% \pm 0.09\%$ \\ %
ResNeXt50~\cite{xie2017aggregated}              & $78.55\% \pm 0.22\%$ & $94.61\% \pm 0.16\%$ & $78.92\% \pm 0.14\%$ & $94.80\% \pm 0.12\%$ \\ %
ResNeXt101~\cite{xie2017aggregated}             & $79.13\% \pm 0.33\%$ & $94.85\% \pm 0.14\%$ & $79.54\% \pm 0.25\%$ & $94.84\% \pm 0.10\%$ \\ %
ResNeXt152~\cite{xie2017aggregated}             & $79.26\% \pm 0.29\%$ & $94.69\% \pm 0.11\%$ & $79.48\% \pm 0.16\%$ & $94.89\% \pm 0.17\%$ \\ %
SE-ResNet18~\cite{hu2018squeeze}                & $76.25\% \pm 0.18\%$ & $93.09\% \pm 0.19\%$ & $76.77\% \pm 0.10\%$ & $93.36\% \pm 0.09\%$ \\ %
SE-ResNet34~\cite{hu2018squeeze}                & $77.85\% \pm 0.19\%$ & $93.88\% \pm 0.15\%$ & $78.20\% \pm 0.16\%$ & $94.13\% \pm 0.21\%$ \\ %
SE-ResNet50~\cite{hu2018squeeze}                & $77.78\% \pm 0.26\%$ & $94.33\% \pm 0.12\%$ & $78.79\% \pm 0.11\%$ & $94.53\% \pm 0.24\%$ \\ %
SE-ResNet101~\cite{hu2018squeeze}               & $77.94\% \pm 0.49\%$ & $94.22\% \pm 0.10\%$ & $79.19\% \pm 0.37\%$ & $94.70\% \pm 0.13\%$ \\ %
SE-ResNet152~\cite{hu2018squeeze}               & $78.10\% \pm 0.47\%$ & $94.46\% \pm 0.13\%$ & $79.35\% \pm 0.27\%$ & $94.73\% \pm 0.15\%$ \\ %
NASNet~\cite{zoph2018learning}                  & $77.76\% \pm 0.19\%$ & $94.26\% \pm 0.28\%$ & $78.17\% \pm 0.11\%$ & $94.35\% \pm 0.21\%$ \\ %
Wide-RN-40-10~\cite{zagoruyko2016wide}          & $78.93\% \pm 0.07\%$ & $94.42\% \pm 0.09\%$ & $79.60\% \pm 0.18\%$ & $94.80\% \pm 0.12\%$ \\ %
StochD-RN-18~\cite{huang2016deep}               & $75.39\% \pm 0.14\%$ & $94.09\% \pm 0.10\%$ & $75.44\% \pm 0.33\%$ & $94.13\% \pm 0.17\%$ \\ %
StochD-RN-34~\cite{huang2016deep}               & $78.03\% \pm 0.33\%$ & $94.81\% \pm 0.08\%$ & $78.16\% \pm 0.39\%$ & $94.97\% \pm 0.10\%$ \\ %
StochD-RN-50~\cite{huang2016deep}               & $77.02\% \pm 0.18\%$ & $94.61\% \pm 0.13\%$ & $77.40\% \pm 0.24\%$ & $94.78\% \pm 0.10\%$ \\ %
StochD-RN-101~\cite{huang2016deep}              & $78.72\% \pm 0.12\%$ & $94.67\% \pm 0.05\%$ & $78.96\% \pm 0.27\%$ & $94.75\% \pm 0.05\%$ \\ %
\midrule
Average (avg.~std)                              & $75.90\%~~(0.26\%)$ & $93.19\%~~(0.15\%)$ & $76.39\%~~(0.24\%)$ & $93.42\%~~(0.14\%)$ \\
\bottomrule
    \end{tabular}
    \caption{Results on CIFAR-100, trained for $200$ epochs, averaged over $5$ seeds including standard deviations.}
    \label{tab:conv-cifar-100-std}
\end{table*}

\begin{table*}[]
    \centering
    \setlength{\tabcolsep}{6pt}
    \small
    \begin{tabular}{l|cc}
\toprule
                                                & \multicolumn{2}{c}{TrAct ($\lambda{=}0.1$, $133$ ep)} \\
Model                                           & Top-1 & Top-5 \\
\midrule
SqueezeNet~\cite{iandola2016squeezenet}         & $70.36\% \pm 0.30\%$ & $91.69\% \pm 0.16\%$ \\ %
MobileNet~\cite{howard2017mobilenets}           & $67.45\% \pm 0.38\%$ & $89.41\% \pm 0.13\%$ \\ %
MobileNetV2~\cite{sandler2018mobilenetv2}       & $68.01\% \pm 0.32\%$ & $90.90\% \pm 0.13\%$ \\ %
ShuffleNet~\cite{zhang2018shufflenet}           & $70.31\% \pm 0.32\%$ & $91.67\% \pm 0.25\%$ \\ %
ShuffleNetV2~\cite{ma2018shufflenet}            & $70.09\% \pm 0.34\%$ & $91.20\% \pm 0.20\%$ \\ %
VGG-11~\cite{simonyan2014very}                  & $69.14\% \pm 0.13\%$ & $88.92\% \pm 0.18\%$ \\ %
VGG-13~\cite{simonyan2014very}                  & $72.53\% \pm 0.26\%$ & $90.81\% \pm 0.12\%$ \\ %
VGG-16~\cite{simonyan2014very}                  & $72.11\% \pm 0.10\%$ & $90.28\% \pm 0.10\%$ \\ %
VGG-19~\cite{simonyan2014very}                  & $70.54\% \pm 0.46\%$ & $88.48\% \pm 0.20\%$ \\ %
DenseNet121~\cite{huang2017densely}             & $79.09\% \pm 0.21\%$ & $94.79\% \pm 0.11\%$ \\ %
DenseNet161~\cite{huang2017densely}             & $80.20\% \pm 0.12\%$ & $95.30\% \pm 0.11\%$ \\ %
DenseNet201~\cite{huang2017densely}             & $79.99\% \pm 0.20\%$ & $95.12\% \pm 0.16\%$ \\ %
GoogLeNet~\cite{szegedy2014going}               & $76.59\% \pm 0.35\%$ & $93.83\% \pm 0.18\%$ \\ %
Inception-v3~\cite{szegedy2016rethinking}       & $78.70\% \pm 0.22\%$ & $94.76\% \pm 0.16\%$ \\ %
Inception-v4~\cite{szegedy2017inception}        & $76.50\% \pm 0.46\%$ & $93.56\% \pm 0.21\%$ \\ %
Inception-RN-v2~\cite{szegedy2017inception}     & $75.15\% \pm 0.24\%$ & $92.99\% \pm 0.29\%$ \\ %
Xception~\cite{chollet2017xception}             & $77.55\% \pm 0.34\%$ & $93.90\% \pm 0.14\%$ \\ %
ResNet18~\cite{he2016deep}                      & $75.86\% \pm 0.20\%$ & $93.07\% \pm 0.07\%$ \\ %
ResNet34~\cite{he2016deep}                      & $77.29\% \pm 0.23\%$ & $93.72\% \pm 0.17\%$ \\ %
ResNet50~\cite{he2016deep}                      & $78.44\% \pm 0.27\%$ & $94.47\% \pm 0.11\%$ \\ %
ResNet101~\cite{he2016deep}                     & $79.20\% \pm 0.17\%$ & $94.77\% \pm 0.11\%$ \\ %
ResNet152~\cite{he2016deep}                     & $79.34\% \pm 0.21\%$ & $94.92\% \pm 0.08\%$ \\ %
ResNeXt50~\cite{xie2017aggregated}              & $78.90\% \pm 0.16\%$ & $94.75\% \pm 0.06\%$ \\ %
ResNeXt101~\cite{xie2017aggregated}             & $79.09\% \pm 0.15\%$ & $94.78\% \pm 0.08\%$ \\ %
ResNeXt152~\cite{xie2017aggregated}             & $78.91\% \pm 0.18\%$ & $94.67\% \pm 0.12\%$ \\ %
SE-ResNet18~\cite{hu2018squeeze}                & $76.51\% \pm 0.43\%$ & $93.29\% \pm 0.16\%$ \\ %
SE-ResNet34~\cite{hu2018squeeze}                & $77.81\% \pm 0.15\%$ & $94.02\% \pm 0.18\%$ \\ %
SE-ResNet50~\cite{hu2018squeeze}                & $78.32\% \pm 0.22\%$ & $94.47\% \pm 0.14\%$ \\ %
SE-ResNet101~\cite{hu2018squeeze}               & $79.07\% \pm 0.12\%$ & $94.79\% \pm 0.31\%$ \\ %
SE-ResNet152~\cite{hu2018squeeze}               & $79.03\% \pm 0.49\%$ & $94.74\% \pm 0.10\%$ \\ %
NASNet~\cite{zoph2018learning}                  & $77.85\% \pm 0.22\%$ & $94.34\% \pm 0.16\%$ \\ %
Wide-RN-40-10~\cite{zagoruyko2016wide}          & $79.37\% \pm 0.25\%$ & $94.72\% \pm 0.07\%$ \\ %
StochD-RN-18~\cite{huang2016deep}               & $74.11\% \pm 0.16\%$ & $93.75\% \pm 0.13\%$ \\ %
StochD-RN-34~\cite{huang2016deep}               & $76.83\% \pm 0.31\%$ & $94.61\% \pm 0.19\%$ \\ %
StochD-RN-50~\cite{huang2016deep}               & $75.87\% \pm 0.29\%$ & $94.28\% \pm 0.18\%$ \\ %
StochD-RN-101~\cite{huang2016deep}              & $77.73\% \pm 0.20\%$ & $94.55\% \pm 0.02\%$ \\ %
\midrule
Average (avg.~std)                              & $75.94\%~~(0.25\%)$ & $93.34\%~~(0.15\%)$ \\

\bottomrule
    \end{tabular}
    \caption{Results on CIFAR-100, trained for $133$ epochs, averaged over $5$ seeds including standard deviations.}
    \label{tab:conv-cifar-100-std-133}
\end{table*}

\end{document}